\documentclass{article}

\usepackage[nonatbib,preprint]{neurips_2025}

\usepackage[utf8]{inputenc} 
\usepackage[T1]{fontenc}
\usepackage{hyperref}
\usepackage{url}
\usepackage{booktabs}
\usepackage{amsfonts}
\usepackage{nicefrac}
\usepackage{microtype}
\usepackage{xcolor}
\usepackage{amsmath}
\usepackage{graphicx,wrapfig}
\usepackage{subcaption}
\usepackage{multirow}
\usepackage{subcaption}
\usepackage{booktabs} 
\usepackage[most]{tcolorbox}
\usepackage{amsmath}
\usepackage{array}
\usepackage{geometry}
\usepackage{enumitem}

\title{FinLMM-R1: Enhancing Financial Reasoning in LMM through Scalable Data and Reward Design}

\author{%
Kai Lan\thanks{Both authors contributed equally to this research.}, \ \ 
Jiayong Zhu$^{*}$, \ \ 
Jiangtong Li, \ \ 
Dawei Cheng, \ \ 
Guang Chen, \ \ 
Changjun Jiang\\
School of Computer Science and Technology, Tongji University \\
\texttt{\{2151042,jiayongz,jiangtongli\}@tongji.edu.cn} \\
}

\begin{document}

\maketitle

\begin{abstract}

Large Multimodal Models~(LMMs) demonstrate significant cross-modal reasoning capabilities.
However, financial applications face challenges due to the lack of high-quality multimodal reasoning datasets and the inefficiency of existing training paradigms for reasoning enhancement.
To address these issues, we propose an integrated framework, FinLMM-R1, combining an automated and scalable pipeline for data construction with enhanced training strategies to improve the multimodal reasoning of LMM.
The \textbf{A}utomated and \textbf{S}calable \textbf{P}ipeline~(\textbf{ASP}) resolves textual-visual misalignment in financial reports through a separate paradigm of question-answer generation and image-question alignment, ensuring data integrity and extraction efficiency.
Through ASP, we collect 89,378 aligned image-question pairs from 23,397 financial reports, covering tasks such as arithmetic reasoning, statistics reasoning, financial explanation, and financial knowledge.
Moreover, we introduce the \textbf{T}hinking with \textbf{A}dversarial \textbf{R}eward in \textbf{LMM}~(\textbf{TAR-LMM}), extending the prior two-stage training framework~\cite{peng2025lmm} with additional reward mechanisms.
In the first stage, we focus on text-only tasks with format and accuracy rewards to guide the model in generating well-structured thinking contents. 
In the second stage, we construct multi-image contrastive samples with additional reward components including image selection, thinking content length, and adversarial reward to jointly optimize the LMM across visual perception, reasoning efficiency, and logical coherence.
Extensive experiments on 7 benchmarks show ASP-derived dataset and training framework significantly improve answer accuracy and reasoning depth over existing reasoning LMMs in both general and financial multimodal contexts.

\end{abstract}

\section{Introduction}

The rapid development of LMMs has driven significant advances in multimodal understanding and reasoning.
Consequently, multimodal learning in the financial domain has also garnered increasing attention.
However, financial data are inherently heterogeneous, encompassing textual data (\emph{e.g.}, reports and news), images (\emph{e.g.}, charts and financial visuals), and tables.
This complexity imposes greater demands on multimodal understanding and reasoning for LMMs than in general domains.

\begin{figure}[t]
  \centering
  \begin{subfigure}[t]{0.523\textwidth}
    \includegraphics[width=\linewidth]{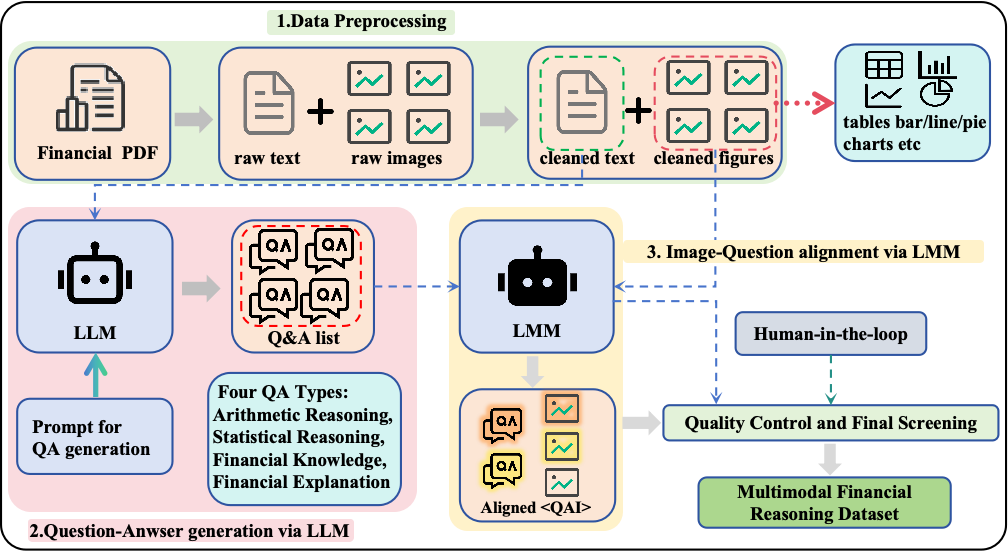}
    \caption{Automated and Scalable Pipeline separates question-answer generation and image-question alignment with LLM and LMM respectively to produce IQA triplets from misaligned textual-visual document.}
    \label{fig:dataset_pipeline}
  \end{subfigure}
  \hfill
  \begin{subfigure}[t]{0.457\textwidth}
    \includegraphics[width=\linewidth]{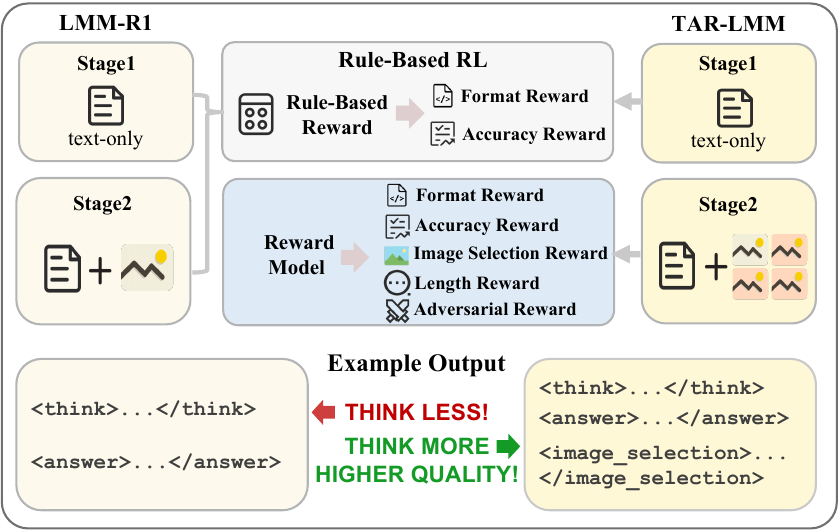}
    \caption{TAR-LMM extends single-image into multi-image with distractors and introduces image selection, length, and adversarial rewards to optimize perception, reasoning efficiency, and quality.}
    \label{fig:model_pipeline}
  \end{subfigure}
  \vspace{-4pt}
  \caption{The illstration of (a) Automated and Scalable Pipeline for multimodal financial data collection and (b) Thinking with Adversarial Reward in LMM~(TAR-LMM) training framework.}
  \label{fig:data}
  \vspace{-10pt}
\end{figure}

Several efforts~\cite{bhatia2024fintral, wang2023finvis} have made notable advances in handling multimodal financial data.
FinTral~\cite{bhatia2024fintral}, based on Mistral-7B, integrates large-scale financial texts, tables, and numerical financial images, which enhances accuracy on multimodal financial tasks through domain-specific pretraining, supervised fine-tuning~(SFT), and direct preference optimization~(DPO).
FinVis-GPT~\cite{wang2023finvis} focuses on financial chart understanding through a diverse dataset comprising multiple chart types and descriptions.
The combination of pretraining and SFT improves its performance in chart description, question answering, and trend prediction, demonstrating the potential of LMMs in finance.

Despite these advances, research on financial multimodal reasoning still faces several critical challenges.
(1) The lack of high-quality open-source multimodal datasets for financial reasoning.
Existing datasets mainly target perception tasks and shallow semantic matching, exhibiting limited logical depth, which limits their ability for cross-modal alignment and reasoning.
Furthermore, most datasets rely on manual annotation and lack scalable, automated pipelines, hindering automated construction and dataset diversity.
(2) The lack of mechanisms to optimize the reasoning process itself.
Most existing LMMs are optimized through supervised objectives, such as instruction tuning based on manually annotated final answers.
Some works also make an attempt on reinforcement learning~(RL). However, the reward functions often focus solely on answer accuracy or thinking format, overlooking intermediate signals such as thinking quality, leading to poor generalization in complex tasks.
Therefore, in financial tasks involving structured information fusion and reasoning, LMMs continue to suffer from limited logical rigor, vision-language coordination, and inference efficiency.

To address the shortage of high-quality data, we propose an \textbf{A}utomated and \textbf{S}calable \textbf{P}ipeline~(\textbf{ASP}) for multimodal financial reasoning dataset construction.
As illustrated in Figure~\ref{fig:dataset_pipeline}, the main challenge in data construction lies in the textual-visual misalignment in financial reports.
Therefore, we adopt a separate paradigm of question-answer generation and image-question alignment to ensure data integrity and extraction efficiency.
In detail, our pipeline first extracts and cleans multimodal content from financial reports.
Then we employ LLM-based question-answer generation and LMM-based image-question alignment to produce image-question-answer~(IQA) triplets.
Finally, human annotators are involved to double-check doubtful questions when three LMMs fail to answer them.

Based on ASP, we construct a large-scale financial multimodal reasoning dataset, covering 23,397 reports from 2016 to 2023 and yielding 89,378 high-quality IQA triplets.
Tasks span arithmetic reasoning, statistical reasoning, financial explanation, and financial knowledge, with visual inputs ranging from price-trend charts and inventory tables to valuation matrices across sectors such as agriculture and property management.
ASP enhances construction efficiency, enables stable, large-scale production of high-quality structured samples, and compared with existing datasets, provides higher automation to underpin structured-reasoning model development.

To further enhance multimodal reasoning, in Figure~\ref{fig:model_pipeline}, we introduce \textbf{T}hinking with \textbf{A}dversarial \textbf{R}eward in \textbf{LMM}~(\textbf{TAR-LMM}), a training framework extending the two-stage LMM-R1~\cite{peng2025lmm}, which improves both reward design and multimodal task formulation.
TAR-LMM employs a two-stage process, whose first stage uses text-only inputs with format and accuracy rewards to learn coherent thinking contents.
In the second stage, we extend single-image inputs into multi-image settings with distractors, and we introduce additional image selection, length, and adversarial rewards to jointly optimize visual perception, reasoning efficiency, and logical quality of LMM.
The adversarial reward employs a BERT model to predict the correlation between the thinking process and the answer correctness. 
In this setup, the BERT serves as a discriminator to determine whether the thinking can lead to a correct answer, while the LMM serves as a generator that attempts to deceive the BERT with reasonable thinking content.
By optimizing the LMM and BERT in turn, the LMM can generate more reasonable thinking processes, while the BERT can better capture flaws within the reasoning.
Compared to the accuracy and format reward, TAR-LMM offers a more robust optimization that yields well-structured thinking content and precisely grounds key visual information.

We evaluate the FinLMM-R1 by benchmarking against LMM-R1 on one text reasoning benchmark, four general multimodal reasoning benchmarks, and two financial multimodal reasinng benchmarks.
Extensive experiments on 7 benchmarks show that the ASP-derived dataset and the training framework significantly improves answer accuracy and reasoning depth over existing reasoning LMMs in both general and financial domain.
Additionally, we conduct ablation studies to evaluate the ASP-derived dataset and each TAR-LMM reward component, and provide insights into our optimization strategies.
Our contribution can be summarized as:
\vspace{-5pt}
\begin{itemize}
    \item We systematically analyze the key challenges in financial multimodal reasoning, including the lack of high-quality datasets and insufficient optimization of the reasoning process.
    
    \item We design an automated and scalable pipeline to construct a large-scale multimodal financial reasoning dataset with charts and structured tables, covering tasks such as arithmetic reasoning, statistical reasoning, financial explanation and financial knowledge.
    
    \item We propose a two-stage multi-reward RL framework for multimodal reasoning, jointly optimizing visual perception, reasoning efficiency, and logical coherence.
    
    \item Extensive experiments on 7 benchmarks show ASP-derived dataset and training framework significantly improve answer accuracy and reasoning depth over existing reasoning LMMs in both general and financial multimodal contexts.
\end{itemize}

\section{Related Work}

\subsection{Multimodal Reasoning for LMMs}
Recent advances in LMMs have been driven by large-scale pre-training and alignment, improving both perceptual and reasoning ability~\cite{radford2021learning,li2023blip,liu2023visual}. 
To extend the reasoning abilities of LMMs, LLM techniques such as in-context learning~(ICL)~\cite{brown2020language} and chain-of-thought~(CoT)~\cite{wei2022chain} prompting have been adapted to multimodal contexts~\cite{alayrac2022flamingo,li2024improving,huang2024multimodal,shao2024visual,chen2024visual}.
MMICL~\cite{zhao2023mmicl} introduces a multimodal ICL mechanism with complex prompts, and DDCoT~\cite{zheng2023ddcot} combines negative space prompting with visual recognition.
Meanwhile, agent-based reasoning approaches have enhanced LMM reasoning capabilities~\cite{lu2023chameleon,yao2023react,tang2023medagents,yang2023mm,qin2023toolllm}.
For example, Liu~\emph{et al.}~\cite{liu2023towards} improve model selection and robustness in multi-step reasoning.
With the introduction of DeepSeek-R1~\cite{guo2025deepseek}, test-time scaling methods based on RL~\cite{guo2025deepseek, zhang2025srpo} have enhanced the reasoning ability of LMM~\cite{wei2025skywork,wang2025visualprm,xing2025echoink,yang2025r1}.
LMM-R1~\cite{peng2025lmm} employs a two-stage rule-based reinforcement learning approach, transitioning from text to multimodal contexts.
However, existing rewards focus mainly on accuracy and format~\cite{guo2025deepseek}, overlooking intermediate signals such as the quality of reasoning, leading to poor generalization in complex tasks.
Therefore, we extend single-image inputs into multi-image settings with distractors, and introduce additional image selection, length, and adversarial rewards to jointly optimize LMM.

\subsection{LLMs and LMMs in Finance}
Open-source LLMs and LMMs (\emph{e.g.}, Qwen~\cite{bai2023qwen}, LLaMA~\cite{grattafiori2024llama}) and proprietary models (\emph{e.g.}, GPT~\cite{brown2020language}, Gemini~\cite{team2023gemini}) have demonstrated promising performance in financial applications.
To align with the unique knowledge system of the financial domain, financial foundation models~\cite{wu2023bloomberggpt, wang2023fingpt, zhang2023xuanyuan, li2023cfgpt, chen2023disc, xie2023pixiu, liu2025fin} have made significant advances. 
Among them, BloombergGPT~\cite{wu2023bloomberggpt} is the first large-scale model dedicated to finance, and FinGPT~\cite{wang2023fingpt} incorporates data-centric fine-tuning for financial tasks.
In the Chinese financial domain, LLMs such as Xuanyuan~\cite{zhang2023xuanyuan}, CFGPT~\cite{li2023cfgpt}, DISC-FinLLM~\cite{chen2023disc}, and PiXiu~\cite{xie2023pixiu} leverage local data for pretraining and supervised fine-tuning~(SFT).
For reasoning models in LLMs, Fin-R1~\cite{liu2025fin} and XuanYuan-FinX1-Preview~\cite{zhang2023xuanyuan} combine SFT and RL to enhance reasoning in financial contexts.
LMMs like FinVis-GPT~\cite{wang2023finvis} and FinTral~\cite{bhatia2024fintral} achieve multimodal performance on par with GPT-4~\cite{achiam2023gpt} while mitigating financial hallucinations through pretraining and SFT on financial charts.
However, reasoning models in multimodal financial domain remain underexplored.
Our work addresses this by developing an integrated framework, FinLMM-R1, combining an automated and scalable pipeline with enhanced training framework to produce reasoning LMM in finance.

\subsection{Benchmarks and Datasets in Finance}
Existing financial benchmarks include FLUE~\cite{shah2022flue}, CFLEB~\cite{lu2023bbt}, and DISC-FinLLM~\cite{chen2023disc} for evaluating financial NLP capabilities, and FinEval~\cite{zhang2023fineval}, InvestorBench~\cite{li2024investorbench}, and FinQA~\cite{chen2021finqa} for financial knowledge and analysis.
Benchmarks integrating financial NLP tasks and broader skill assessments include FinBen~\cite{xie2024finben}, SuperCLUE-Fin~\cite{xu2024superclue}, FinanceIQ~\cite{zhang2023xuanyuan}, and OpenFinData~\cite{openfindata}.
Benchmarks such as FAMMA~\cite{xue2025famma} and MME-Finance~\cite{gan2024mme} are introduced to evaluate multimodal capabilities like question answering based on financial charts and tables.
There are few publicly available financial datasets.
Among them YiZhao-FinDataSet~\cite{yizhao_findataset} and FinCorpus~\cite{zhang2023xuanyuan} are designed for pretraining. 
Most datasets designed for financial tasks, such as DISC-Fin-SFT~\cite{chen2023disc}, are not publicly available.
To address the lack of multimodal financial data for reasoning task, we develop an automated and scalable data construction pipeline from financial reports. 
This results in a large-scale dataset comprising 23,397 reports and 89,378 IQA triplets for training and evaluation.

\section{Automated and Scalable Data Construction}

\begin{figure*}[t]
\centering
\includegraphics[width=0.95\textwidth]{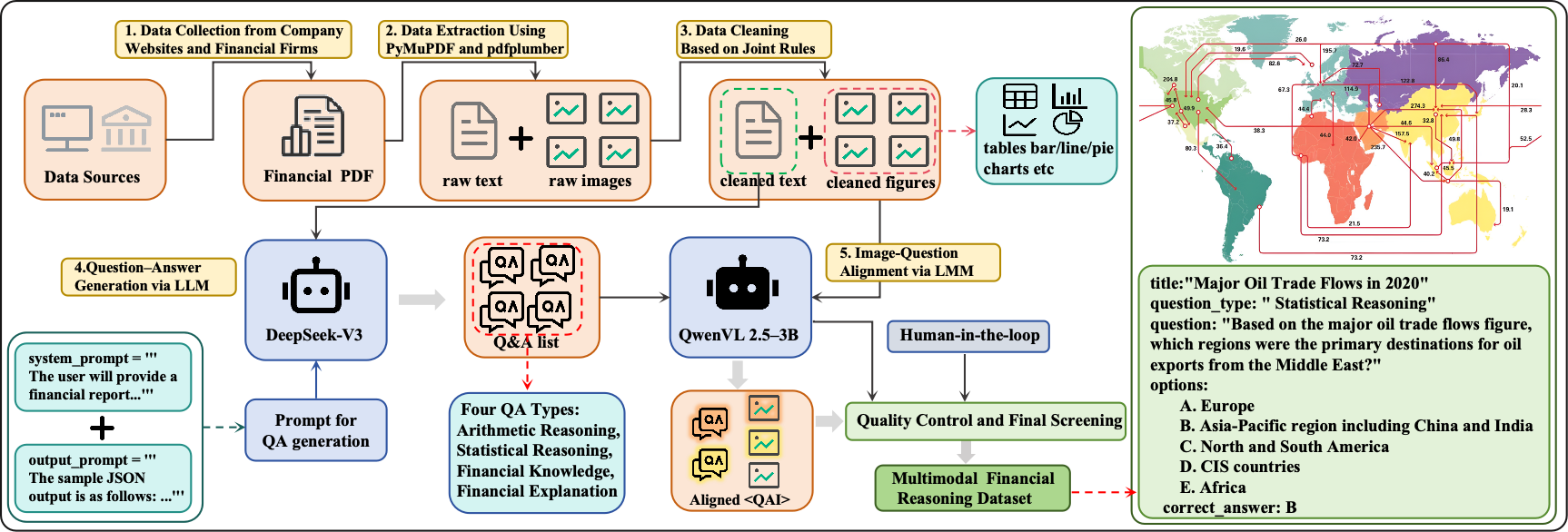}
\caption{The illustration shows the ASP for multimodal financial data collection. We first extract and clean multimodal content from financial reports. We then employ LLM-based question-answer generation and LMM-based image-question alignment to generate IQA triplets separately. Finally, human annotators are involved to double-check doubtful questions.}
\label{fig:ASP-framework}
\end{figure*}

\subsection{Data Collection and Extraction}
We collect financial reports of Chinese listed companies from 2016 to 2023, which are sourced from official company websites and third-party financial platforms, where only PDF documents is retained for format consistency.
During content extraction, we use PyMuPDF and pdfplumber to retrieve two key elements from each PDF: (1) all machine-readable text, and (2) embedded images and charts.
The extracted text is stored in JSON format, and images are grouped and saved by document.

\subsection{Data Cleaning}
To ensure the quality of our multimodal financial dataset, we apply a series of rule-based strategies to clean and refine the extracted content.
For text, we first remove irrelevant parts such as disclaimers, addresses, and references based on keyword rules, and merge fragmented sentence segments.
We further apply page-level filtering using features such as text length, character ratio, and punctuation distribution.
We also correct misidentified chart descriptions and merge directory pages to ensure semantic completeness.
For images, we remove low-resolution, small, or irrelevant figures, and apply image similarity metrics to eliminate duplicates (\emph{e.g.}, corporate logos or repeated icons).
To identify valid images, we examine whether textual reference marks, such as ``Chart x'', ``Figure x'', and ``Table x'', appear in the corresponding document.
PDFs with extracted images covering less than 70\% of the total charts are discarded to ensure reliable image-text correspondence.

\subsection{Separated Question-Answer Generation and Image-Question Alignment}

Financial reports contain in-depth analysis of charts and tables.
However, imperfect typesetting often causes misalignment between text and charts.
Therefore, we propose a separate paradigm of question-answer generation and image-question alignment to ensure data integrity and efficiency.

\textbf{Question-Answer Generation via LLM.}
In this step, we employ DeepSeek-V3~\cite{liu2024deepseek} to generate chart-relevant multiple-choice questions based on the text of each document.
To this end, we design structured prompts to guide the LLM in locating chart descriptions and generating corresponding question-answer pairs.
Each question consists of five answer options, a correct answer, and a brief explanation to facilitate human checking.
Following the question categories from existing financial benchmarks~\cite{zhang2023fineval, li2024investorbench, openfindata, gan2024mme}, we design the questions to cover four categories: \textbf{Arithmetic Reasoning, Statistical Reasoning, Financial Explanation,} and \textbf{Financial Knowledge}. The former three categories focus on reasoning ability, while the last emphasizes financial knowledge, aiming to enhance fine-grained reasoning and comprehension capabilities in multimodal financial reasoning.

\textbf{Image-Question Alignment via LMM.}
In this step, we use the locally deployed QwenVL 2.5-3B~\cite{bai2025qwen2} to evaluate the semantic relevance between questions and candidate images within the same document.
To ensure a one-to-one correspondence, we retain only image-question pairs with a clear and unique match; otherwise, the questions are discarded to avoid ambiguous supervision.
All images from the same document are retained for downstream image selection tasks.
After constructing the IQA triplets, we randomly sample 100 documents with 384 IQAs for manual verification.
Among them, 358 IQAs are evaluated as semantically aligned and answerable by three finance graduate annotators.
This approach improves alignment quality and supports structurally reliable training data.

\subsection{Quality Control and Final Filtering}
During manual verification in the last step, we notice that semantically unaligned and unanswerable IQAs often receive uncertain answers from LMMs.
Therefore, to further ensure IQA quality, we perform three rounds of inference using QwenVL 2.5-3B~\cite{bai2025qwen2}, where LMM answers the question based on the corresponding image.
IQAs that receive three different answers from LMM are selected for additional manual review.
Over 6,500 IQAs are manually examined by three finance graduate student annotators.
Only IQAs evaluated as semantically aligned and answerable by all three annotators are retained in the dataset.
In the final filtering step, we retain reports containing 2 to 10 valid images to balance the image diversity and computational cost.
The dataset consists of 23,397 financial reports from 2016 to 2023, yielding 89,378 IQAs, with additional statistics presented in Supplementary Materials, which offers semantically coherent, and moderately challenging IQA triplets, enabling training for multimodal financial understanding and reasoning.

\section{Methodology}

\begin{figure*}[t]
\centering
\includegraphics[width=0.96\textwidth]{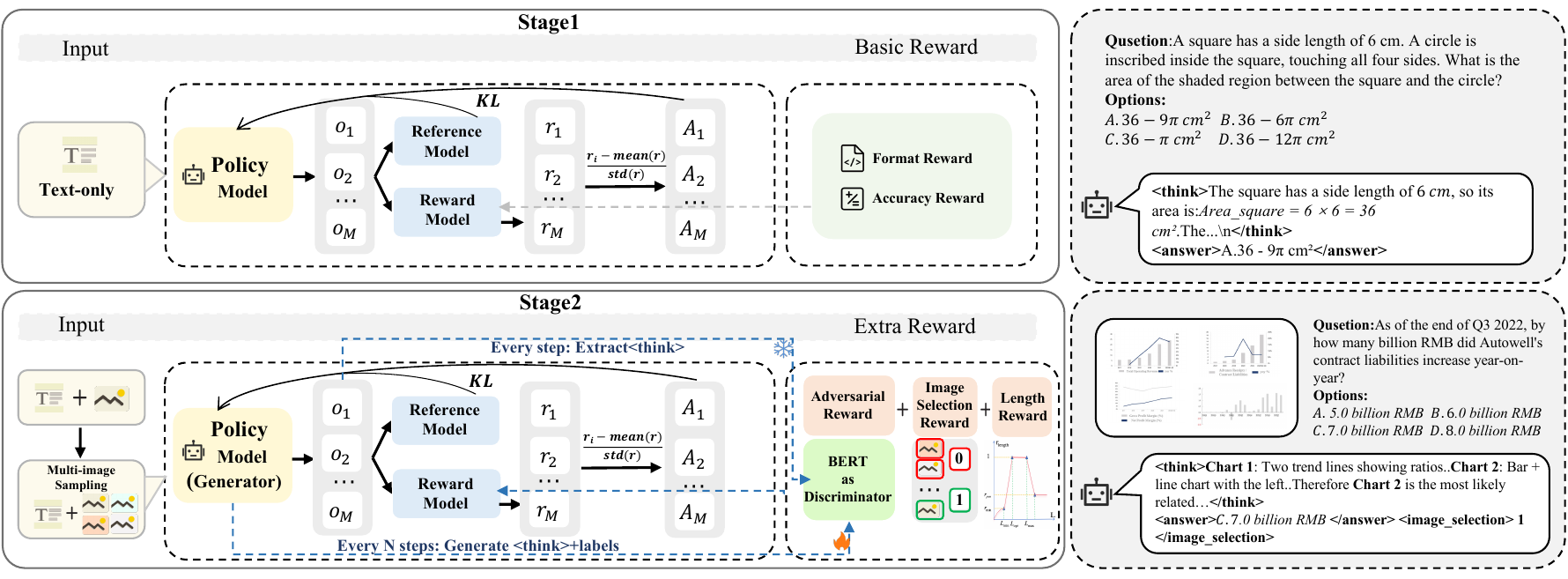}
\caption{The illustration shows the TAR-LMM training framework. In the first stage, we focus on text-only tasks with format and accuracy rewards, enhancing LMM reasoning ability. In the second stage, we construct multi-image contrastive samples with additional three rewards.}
\label{fig:TARLMM-framework}
\end{figure*}

To enhance LMM performance on multimodal reasoning tasks, we extend the two-stage training framework of LMM-R1 with a more efficient RL optimization algorithm in Figure~\ref{fig:TARLMM-framework}.

\subsection{Reinforcement Learning for LMMs}

To enhance LMM performance on mathematical and multimodal reasoning tasks, we adopt the \textit{Group Relative Policy Optimization}~(GRPO) algorithm~\cite{shao2024deepseekmath} to optimize the policy LMM.
GRPO addresses two key limitations of the Proximal Policy Optimization~(PPO): (1) PPO typically requires training an additional value model, increasing computational and memory overhead; (2) only the final token in output is scored by reward model, making it difficult to accurately estimate values for other tokens.

GRPO avoids the need for explicit value function estimation by computing advantages based on relative rewards within a sampled group. 
Specifically, for each input query \( q \), a set of outputs \( \{o_1, o_2, ..., o_G\} \) is sampled using the old policy \( \pi_{\text{old}} \), and each sample is scored by a reward model to obtain rewards \( \{s_1, s_2, ..., s_G\} \).
These rewards are normalized within the group and used to compute shared advantages for guiding policy optimization.
The GRPO objective can be formalized as
\begin{equation}
\mathcal{J}_{\text{GRPO}}(\theta)\! = \!\mathbb{E}_{q, \{o_i\}} \!\left[\!  
\frac{1}{G} \sum_{i=1}^G \frac{1}{|o_i|} \sum_{t=1}^{|o_i|}
\!\left\{\!
\min \!\left(\! r_{i,t} \hat{A}_{i,t}, \text{clip}(\!r_{i,t}, 1\!-\!\varepsilon, 1\!+\!\varepsilon\!) \hat{A}_{i,t} \!\right)\!
\!-\! \beta \mathbb{D}_{\text{KL}}[\pi_\theta \| \pi_{\text{ref}}]
\!\right\}\!
\!\right]\!
\end{equation},
where \( \hat{A}_{i,t} \) denotes the advantage computed from normalized group rewards, \( \beta \) is the coefficient for the Kullback-Leibler~(KL) divergence regularization term, and \( \pi_{\text{ref}} \) is the reference policy model.

Unlike PPO, GRPO incorporates the KL divergence into the loss function instead of modifying the reward, thereby maintaining stability in advantage estimation.
Moreover, since the reward model is designed to assess the relative quality among outputs, the intra-group comparison mechanism of GRPO aligns closely with the design of the reward model.
Our TAR-LMM framework is built on the GRPO algorithm by adopting a two-stage strategy to improve multimodal reasoning ability of LMMs.

\subsection{Stage-1: Text-only Reasoning with Rule-Based Rewards}
\label{sec:stage1}

In the first training stage, we follow LMM-R1 using a purely textual dataset and apply rule-based RL to enhance the reasoning ability of LMM.
Specifically, we adopt the GRPO algorithm for optimization, and the reward function in this stage comprises two reward components.

\textbf{Format Reward:}
This reward checks whether the model output satisfies structural formatting requirements. 
The reasoning process must be enclosed within the \texttt{<think></think>} tags, and the final answer must appear within the \texttt{<answer></answer>} tags. 
Only outputs with fully correct format receive a reward of 1; otherwise, the reward is 0.

\textbf{Accuracy Reward:}
This reward checks whether the model correctly answer the question.
We compare the answer enclosed in the \texttt{<answer></answer>} with the reference answer. 
Only answers that are exact equivalent to the reference answer receive a reward of 1; otherwise, the reward is 0.

\textbf{Total Reward:}
These two reward components are combined in a weighted manner to form the total reward function for Stage-1, which can be formulated as:
\begin{equation}
\mathcal{R}_{\text{stage-1}} = \lambda \cdot r_{\text{accuracy}} + (1 - \lambda) \cdot r_{\text{format}},
\end{equation}
where \( \lambda \) is a weighting coefficient balancing accuracy and format rewards.

\subsection{Stage-2: Multimodal Reasoning with Extended Rewards}
\label{sec:stage2}

In the second training stage, we introduce multimodal datasets with three extra rewards, \textit{length reward}, \textit{adversarial reward}, and \textit{image selection reward} further enhance multimodal reasoning.

\textbf{Length Reward.}
This reward is based on a systematic study between reasoning length and reasoning quality.
Consistent with findings from Yang~\emph{et al.}~\cite{yang2025towards}, we observe that the correlation between reasoning length and answer accuracy is non-linear. 
Simply extending the reasoning content does not necessarily improve performance and may cause redundancy.
To address this issue, we design a piecewise reward function that encourages generation of informative and appropriately long reasoning processes.
Specifically, the reward increases sub-linearly in short regions, grows linearly in a moderate range, remains constant in an optimal length window, and slightly declines for very long outputs with a relatively high score.
The length reward \(r_{\mathrm{length}}(L)\) is defined by four regimes: when \(L < L_{\min}\), it follows a square-root schedule, \(r_{\min}\sqrt{L/L_{\min}}\), reaching \(r_{\min}\) at \(L = L_{\min}\); for \(L_{\min}\le L\le L_{\mathrm{opt}}\), it increases linearly via 
\(
r_{\min} + (1 - r_{\min})\frac{L - L_{\min}}{L_{\mathrm{opt}} - L_{\min}},
\)
attaining its maximum of 1 at \(L = L_{\mathrm{opt}}\); it remains constant at 1 for \(L_{\mathrm{opt}}<L\le L_{\max}\); and for \(L > L_{\max}\), it decreases linearly according to 
\(
\max\!\bigl(r_{\mathrm{pen}},\,1 - \gamma\,\tfrac{L - L_{\max}}{L_{\max}}\bigr),
\)
ensuring it never falls below \(r_{\mathrm{pen}}\),
where \( L \) is the reasoning length; \( L_{\min}, L_{\text{opt}}, L_{\max} \) are the minimum, optimal, and maximum target lengths; \( r_{\min} \) is the base reward for short lengths; \( r_{\text{pen}} \) is the penalty floor for long outputs; and \( \gamma \) is the decay factor.

\textbf{Adversarial Reward.}
High-quality reasoning is essential for reasoning LMMs.
However, length rewards can cause the model to generate shallow, verbose output. 
Existing reward designs often focus on the answer, overlooking the logical quality of the reasoning process itself.
To address this, inspired by GAN~\cite{goodfellow2014generative}, we propose an adversarial reward, employing a BERT model to predict answer correctness based on the reasoning process.
In this setup, the BERT serves as a discriminator to assess whether reasoning leads to correct answers, while the LMM acts as a generator attempting to produce plausible reasoning that deceives the BERT.
By optimizing the LMM and BERT in turn, the LMM can generate more reasonable thinking content, while the BERT can better capture flaws in thinking.

The training process of the discriminator proceeds as follows: 
1) At every fixed interval of \( N \) steps, the policy LMM generates a batch of reasoning samples and records the corresponding thinking content.
2) The correctness of each final answer is used as a binary label, where \( 1 \) indicates correct and \( 0 \) indicates incorrect.
3) These labeled samples are used to fine-tune the BERT-based discriminator, enabling it to better assess the quality of reasoning.
We exploit the hidden states from the last BERT layer by concatenating the max-pooled and mean-pooled representations for correctness prediction.
During reinforcement learning, the discriminator receives thinking content $T$ (\emph{i.e.}, the content between \texttt{<think></think>}) and outputs a confidence score in the range $[0, 1]$ as the adversarial reward:
\(r_{\text{adv}} = D_{\phi}(T),\)
where $T$ is the generated reasoning and $D_{\phi}$ is the parameterized discriminator.
This adversarial mechanism guides the policy LMM to iteratively refine its reasoning process by leveraging feedback signals from the discriminator, thereby improving its overall reasoning capability.

\textbf{Image Selection Reward.}
This rewards aims to enhance the visual perception ability of LMMs by encouraging them to identify the most relevant image among a set of candidates.
For each question, we construct samples containing one correct image and 2 to 4 distractors, requiring the model to select the most relevant image.
If the LMM correctly selects the image, the reward, \(r_{\text{img}}\), is 1; otherwise, the reward is 0.
To implement this, we prompt the model to select an image index and wrap its response within the \texttt{<image\_selection></image\_selection>}, making it explicit and evaluable.

\textbf{Total Reward.}
We combine the three new rewards with the original two rewards as:
\begin{equation}
\mathcal{R}_{\text{stage-2}} = \lambda_0 \cdot r_{\text{format}} + \lambda_1 \cdot r_{\text{accuracy}} + \lambda_2 \cdot r_{\text{length}} + \lambda_3 \cdot r_{\text{adv}} + \lambda_4 \cdot r_{\text{img}},
\end{equation}
where \( \lambda_0, \lambda_1, \lambda_2, \lambda_3, \lambda_4 \in [0, 1] \) are coefficients satisfying
\(
\sum_{i=0}^{4} \lambda_i = 1,
\)
balancing different rewards.

\section{Experiments}

\subsection{Experimental Setup}

\textbf{Training Dataset.}
We adopt a two-stage training strategy to enhance LMM reasoning ability.
Following a similar setting in LMM-R1~\cite{peng2025lmm}, we use the \textbf{DeepScaleR-Preview-Dataset} \cite{luo2025deepscaler}for the first stage.
This dataset consists of approximately 40k question-answer pairs as text.
In the second stage, to ensure fair comparison, we adopt the same dataset used in LMM-R1, \emph{i.e.}, \textbf{Verifiable Multimodal-65K~(VerMulti)}.
Furthermore, to verify the effectiveness of our ASP-derived multimodal financial dataset, we randomly sample 40k IQA triplets from 2016-2022 as FinData and merge them with VerMulti, forming a combined dataset for training in the second stage.
To enable the image selection reward being optimized during second training stage, for each sample in VerMulti, we select images corresponding to questions with the lowest possible textual similarity to the current question as distractors; for FinData, we keep the images from the same document as distractors.

\textbf{Evaluation Dataset.}
We evaluate our method on both general-purpose and finance-specific multimodal reasoning tasks.
For general reasoning, we use four multimodal datasets, \textbf{Math Vision}-test~\cite{wang2024measuring}, \textbf{Math Verse}-testmini~\cite{zhang2024mathverse}, \textbf{MMStar}~\cite{chen2024we}, and \textbf{Math Vista}-testmini~\cite{lu2023mathvista}, along with a text reasoning dataset, \textbf{GPQA}-main~\cite{rein2024gpqa}.
For financial reasoning, we construct a test set from the 2023 ASP-derived dataset and employ three finance graduate student annotators to assess alignment and answerability, resulting in 3823 IQA triplets across four categories: \textit{Arithmetic Reasoning}~\textbf{(AR)}, \textit{Statistical Reasoning}~\textbf{(SR)}, \textit{Financial Explanation}~\textbf{(FE)}, and \textit{Financial Knowledge}~\textbf{(FK)}.
We also evaluate on the multimodal subset of \textbf{FAMMA}-basic~\cite{xue2025fammabenchmarkfinancialdomain} as an additional financial benchmark.

\textbf{Compared Methods.} We compare our method with three baseline LMMs: R1-Onevision-3B~\cite{yang2025r1}, InternVL2.5-8B~\cite{chen2024expanding}, and MiniCPM-V2.6-8B~\cite{yao2024minicpm}.
Moreover, since our method is inspired by LMM-R1~\cite{peng2025lmm} and built on Qwen2.5-VL-3B~\cite{bai2025qwen2}, we also compare with them, where their results are copied from LMM-R1~\cite{peng2025lmm}.
The results of stage-1 training (FinLMM-R1 (S1)), stage-2 training with VerMulti (FinLMM-R1 (S2)), and stage-2 training with VerMulti and FinData (FinLMM-R1 (S2) + FinData) are used to validate the effectiveness of our dataset and training framework.
Further details about the experiments are listed in the Supplementary Material.

\subsection{Main Results}

\subsubsection{General Multimodal Reasoning}
In Table~\ref{tab:exp:gmm}, we compare FinLMM-R1(S2) and FinLMM-R1(S2)+FinData on general multimodal reasoning tasks.
On the text reasoning task, we observe that stage-2 multimodal training (FinLMM-R1(S2)) significantly improves performance over stage-1 text-only training by more than 3\%, indicating the complementarity between text and multimodal reasoning.
On the multimodal reasoning tasks, our model FinLMM-R1(S2) outperforms the baseline LMM-R1 across all four datasets, demonstrating the advantages of our additional three rewards.
Comparing improvements across the four datasets, we observe that MM-Star and MathVista~(2-3\%) achieve greater gains than MathVision and MathVerse~(1\%).
Given that MathVision and MathVerse are more difficult, the difference may be attributed to the adversarial rewards, where the discriminator BERT may have difficulty assessing the reasoning correctness for ``hard thinking''.
Moreover, we observe that our model with FinData, FinLMM-R1(S2)+FinData, has similar performance to FinLMM-R1(S2), indicating that financial multimodal reasoning data provides limited benefit in general domains.
Overall, our training framework with three additional rewards effectively improves the reasoning ability of reasoning LMM in general domain, outperforming existing baselines in both text and multimodal reasoning settings.

\begin{table}[t]
\caption{Experimental results on general multimodal reasoning tasks. Best results are in \textbf{bolded}.}
\label{tab:exp:gmm}
\centering
\setlength\tabcolsep{16pt}
\resizebox{0.95\textwidth}{!}{
\begin{tabular}{l|c|cccc|c}
\toprule
\multirow{2}{*}{Model}            & \multicolumn{1}{c|}{Text-only} & \multicolumn{4}{c|}{Multimodal}  & \multirow{2}{*}{Average} \\ \addlinespace[2pt]\cline{2-6} \addlinespace[2pt]
                      & GPQA         & MathVision   & MathVerse    & MM-Star      & MathVista    &        \\ \midrule
R1-Onevision-3B       & -            & 23.70        & 38.60        & -            & -            & -      \\
InternVL2.5-8B        & -            & -            & 35.60        & -            & 64.50        & -      \\
MiniCPM-V2.6-8B       & -            & 23.40        & -            & 57.50            & 60.80        & -      \\ \midrule
Qwen2.5-VL-3B CoT     & 30.30        & 23.59        & 34.64        & 51.40        & 60.70        & 40.13  \\
LMM-R1                & 38.84        & 26.35        & 41.55        & 58.03        & 63.20        & 45.59  \\
FinLMM-R1(S1)         & 36.82        & 25.63        & 38.39        & 54.99        & 61.52        & 43.47  \\
FinLMM-R1(S2)         &\textbf{41.52}&\textbf{27.05}&\textbf{42.29}& 61.13        & 65.20        & 47.43  \\
FinLMM-R1(S2)+FinData & 41.29        & 26.99        & 42.21        &\textbf{61.33}&\textbf{65.40}& \textbf{47.45}    \\ \bottomrule
\end{tabular}
}
\vspace{-10pt}
\end{table}

\begin{table}[t]
\caption{Experimental results on financial multimodal reasoning tasks. Best results are in \textbf{bolded}.}
\label{tab:exp:fmm}
\centering
\setlength\tabcolsep{14pt}
\resizebox{0.95\textwidth}{!}{
\begin{tabular}{l|ccccc|cccc}
\toprule
\multirow{2}{*}{Model} & \multicolumn{5}{c|}{Ours}              & \multicolumn{4}{c}{FAMMA}     \\ \addlinespace[2pt]\cline{2-10} \addlinespace[2pt]
                       & AR    & SR    & FE    & FK    & Tot.  & Easy  & Middle & Hard  & Tot.  \\ \midrule
Qwen2.5-VL-3B CoT      & 39.21 & 33.16 & 41.93 & 62.10 & 44.58 & 27.24 & 12.74  & 10.05 & 18.84 \\
LMM-R1                 & 46.13 & 42.11 & 51.85 & 70.57 & 53.01 & 35.89 & 25.47  & 12.17 & 26.59 \\
LMM-R1+FinData         & 47.08 & 51.58 & 49.60 & 72.78 & 53.87 & 36.78 & 26.11  & 13.23 & 27.46 \\
FinLMM-R1(S1)          & 44.46 & 41.05 & 49.21 & 67.34 & 50.73 & 34.30 & 22.21  & 13.19 & 25.35 \\
FinLMM-R1(S2)          & 49.70 & 46.84 & 52.51 & 70.97 & 55.02 & 35.22 & 23.57  & 17.99 & 27.49 \\
FinLMM-R1(S2)+FinData  & \textbf{53.52} & \textbf{52.11} & \textbf{57.67} & \textbf{74.60} & \textbf{59.36} & \textbf{36.26} & \textbf{24.84}  & \textbf{18.52} & \textbf{28.44} \\ \bottomrule
\end{tabular}
}
\vspace{-10pt}
\end{table}

\subsubsection{Financial Multimodal Reasoning}
In Table~\ref{tab:exp:fmm}, we compare FinLMM-R1(S2) and FinLMM-R1(S2)+FinData on financial multimodal reasoning.
On the ASP-derived dataset, two-stage training boosts the reasoning performance of FinLMM-R1(S2) over LMM-R1, achieving over 2\% improvement and demonstrating the effectiveness of our additional rewards.
Moreover, with FinData, we achieve further improvements across all tasks, resulting in an average improvement of 4\%.
Furthermore, comparing improvements between financial reasoning and knowledge tasks, we find that financial reasoning benefits more, as the two-stage RL training better supports reasoning exploration.
On the FAMMA dataset, our models, FinLMM-R1(S2) and FinLMM-R1(S2)+FinData, outperform the baseline methods, LMM-R1 and LMM-R1+FinData, particularly at the ``Hard'' level, highlighting the effectiveness of our adversarial reward in enhancing reasoning quality and accuracy.
Overall, our framework with three additional rewards and ASP-derived FinData effectively improves the reasoning ability of baseline LMMs in financial multimodal reasoning, validating the potential of data ASP for future advancements.

\subsection{Albation Study}
\textbf{Effect of Different Rewards:}
In Table~\ref{tab:exp:abl}, we study the effects of three additional rewards.
For MathVision and MathVerse, we observe that the length and image selection rewards have a significant impact, possibly because the discriminator BERT struggles to assess reasoning correctness for ``hard questions''. 
For MM-Star and MathVista, we find that the adversarial reward contributes the most, demonstrating its effectiveness in enhancing reasoning quality. 
In financial domain, all three rewards help improve the accuracy, validating their effectiveness for multimodal financial reasoning tasks.

\begin{table}[t]
\caption{Ablation study on different rewards. Most contributed rewards are highlighted with \underline{underline}.}
\label{tab:exp:abl}
\centering
\setlength\tabcolsep{13pt}
\resizebox{0.95\textwidth}{!}{
\begin{tabular}{l|cccc|cc}
\toprule
\multirow{2}{*}{Model} & \multicolumn{4}{c|}{General}                 & \multicolumn{2}{c}{Financial} \\ \addlinespace[2pt]\cline{2-7} \addlinespace[2pt]
                       & MathVision & MathVerse & MM-Star & MathVista & Ours      & FAMMA         \\ \midrule
LMM-R1                 &  26.35     & 41.55     & 58.03   & 63.20     & 53.01     & 26.59         \\
FinLMM-R1(S1)          &  25.63     & 38.39     & 54.20   & 61.52     & 50.73     & 25.35         \\
FinLMM-R1(S2)          &  27.05     & 42.29     & 61.13   & 65.20     & 55.02     & 27.49         \\
FinLMM-R1(S2) w/o LR   &  \underline{26.58}     & 41.82     & 60.73   & 65.00     & 53.35         & 26.53              \\
FinLMM-R1(S2) w/o ISR  &  26.62     & \underline{41.72}     & 60.60   & 64.40     & 53.27         & 26.44              \\
FinLMM-R1(S2) w/o AR   &  26.69     & 41.90     & \underline{59.07}   & \underline{64.20}     & \underline{52.83}         & \underline{25.84}              \\ \bottomrule
\end{tabular}
}
\vspace{-10pt}
\end{table}

\textbf{Effect of Financial Dataset:} To validate the effectiveness of our ASP-derived financial dataset, we train stage-2 with 20\%, 40\%, 60\%, 80\%, and 100\% of FinData and evaluate performance on our financial dataset, as shown in Table~\ref{tab:abl:data}. 
As the proportion of FinData increases from 0\% to 100\%, accuracy consistently improves, demonstrating the scalability the data construction pipeline.

\begin{table}[t]
\caption{The effect of our ASP-derived FinData.}
\label{tab:abl:data}
\centering
\resizebox{0.55\textwidth}{!}{
\begin{tabular}{lllllll}
\toprule
Data Utilization & 0\%   & 20\%  & 40\%  & 60\%  & 80\%  & 100\% \\ \midrule
Ours             & 55.02 & 55.92 & 56.74 & 57.53 & 58.90 & 59.36 \\ \bottomrule
\end{tabular}
}
\vspace{-12pt}
\end{table}

\subsection{Case Study}
\begin{figure*}[h]
\centering
\includegraphics[width=\textwidth]{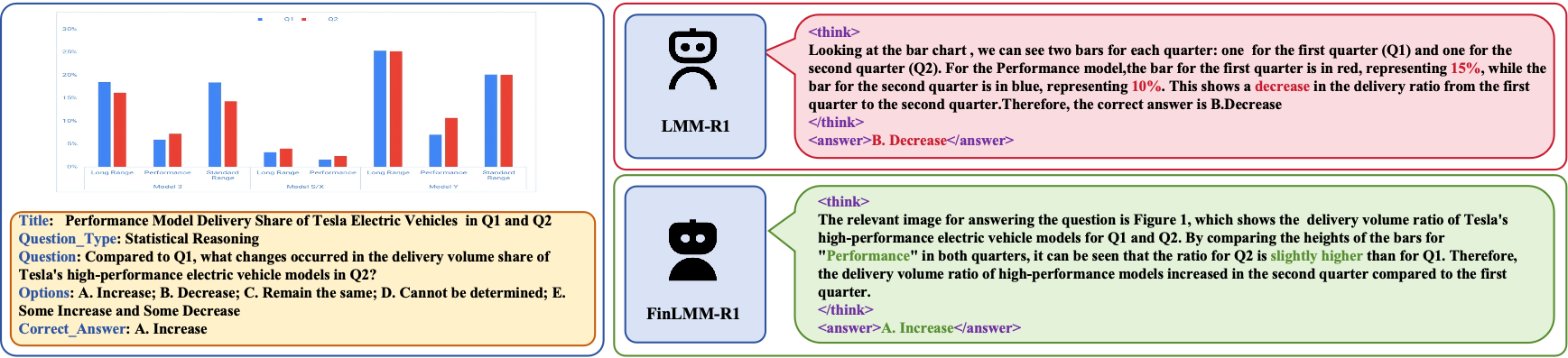}
\caption{Qualitative example showcasing how our FinLMM-R1 framework achieves successful multimodal financial reasoning. Best viewed when zoomed in.}
\label{fig:case_study}
\end{figure*}
In Figure~\ref{fig:case_study}, we provide an example from our financial dataset demonstrating the multimodal financial reasoning ability of FinLMM-R1 framework.
In the example, the image contains information about the ``Performance Model Delivery Share of Tesla Electric Vehicles in Q1 and Q2''.
When asked about the ``delivery volume share of Tesla's high-performance electric vehicle'', the LMM needs to compare the ``performance'' of different Tesla vehicles.
However, LMM-R1 only captures the volume share of the long-range Model-3, indicating its limitations in both visual perception and logical reasoning.
Compared with LMM-R1, FinLMM-R1 can precisely identify the ``performance'' attribute and compare the bar heights across three types of Tesla electric vehicles, reflecting superior abilities in financial concept understanding, visual perception, and logical reasoning.
This case highlights the effectiveness of FinLMM-R1 in addressing complex multimodal financial reasoning challenges.

\section{Conclusion}

In this work, we propose an \textbf{A}utomatic and \textbf{S}calable \textbf{P}ipeline~(\textbf{ASP}) for multimodal financial data construction and a \textbf{T}hink with \textbf{A}dversarial \textbf{R}eward in LMM~(\textbf{TAR-LMM}) training framework for enhancing the reasoning capabilities of LMMs in both general and financial domains.
ASP incorporates a separate paradigm of question-answer generation and image-question alignment to address textual-visual misalignment in financial reports.
TAR-LMM focuses on text-only reasoning tasks with accuracy and format rewards in the first stage to guide the model in developing basic reasoning ability.
In the second stage, we construct multi-image contrastive samples with additional image selection, reasoning length, and adversarial reward, to jointly optimize the LMM across visual perception, reasoning efficiency, and logical coherence.
Extensive experiments across 7 multimodal reasoning benchmarks demonstrate that the data from ASP and the TAR-LMM training framework significantly boost the reasoning performance of LMMs in both general and financial domains.
Overall, this work focuses on addressing data limitations and overlooked reasoning quality, paving the way for future research on LMMs from shadow perception to deep reasoning.

{
\small
\bibliographystyle{neurips_2025}
\bibliography{neurips_2025}

\begin{thebibliography}{10}

\bibitem{peng2025lmm}
Peng, Y., G.~Zhang, M.~Zhang, et~al.
\newblock Lmm-r1: Empowering 3b lmms with strong reasoning abilities through
  two-stage rule-based rl.
\newblock \emph{arXiv preprint arXiv:2503.07536}, 2025.

\bibitem{bhatia2024fintral}
Bhatia, G., E.~M.~B. Nagoudi, H.~Cavusoglu, et~al.
\newblock Fintral: A family of gpt-4 level multimodal financial large language
  models.
\newblock In \emph{ACL Findings}. 2024.

\bibitem{wang2023finvis}
Wang, Z., Y.~Li, J.~Wu, et~al.
\newblock Finvis-gpt: A multimodal large language model for financial chart
  analysis.
\newblock In \emph{FinLLM@IJCAI}. 2023.

\bibitem{radford2021learning}
Radford, A., J.~W. Kim, C.~Hallacy, et~al.
\newblock Learning transferable visual models from natural language
  supervision.
\newblock In \emph{ICML}, pages 8748--8763. 2021.

\bibitem{li2023blip}
Li, J., D.~Li, S.~Savarese, et~al.
\newblock Blip-2: Bootstrapping language-image pre-training with frozen image
  encoders and large language models.
\newblock In \emph{ICML}, pages 19730--19742. 2023.

\bibitem{liu2023visual}
Liu, H., C.~Li, Q.~Wu, et~al.
\newblock Visual instruction tuning.
\newblock In \emph{NeurIPS}, pages 34892--34916. 2023.

\bibitem{brown2020language}
Brown, T., B.~Mann, N.~Ryder, et~al.
\newblock Language models are few-shot learners.
\newblock In \emph{NeurIPS}, vol.~33, pages 1877--1901. 2020.

\bibitem{wei2022chain}
Wei, J., X.~Wang, D.~Schuurmans, et~al.
\newblock Chain-of-thought prompting elicits reasoning in large language
  models.
\newblock In \emph{NeurIPS}, pages 24824--24837. 2022.

\bibitem{alayrac2022flamingo}
Alayrac, J.-B., J.~Donahue, P.~Luc, et~al.
\newblock Flamingo: a visual language model for few-shot learning.
\newblock In \emph{NeurIPS}, pages 23716--23736. 2022.

\bibitem{li2024improving}
Li, W., H.~Fan, Y.~Wong, et~al.
\newblock Improving context understanding in multimodal large language models
  via multimodal composition learning.
\newblock In \emph{ICML}. 2024.

\bibitem{huang2024multimodal}
Huang, B., C.~Mitra, L.~Karlinsky, et~al.
\newblock Multimodal task vectors enable many-shot multimodal in-context
  learning.
\newblock In \emph{NeurIPS}, pages 22124--22153. 2024.

\bibitem{shao2024visual}
Shao, H., S.~Qian, H.~Xiao, et~al.
\newblock Visual cot: Advancing multi-modal language models with a
  comprehensive dataset and benchmark for chain-of-thought reasoning.
\newblock In \emph{NeurIPS}, pages 8612--8642. 2024.

\bibitem{chen2024visual}
Chen, Z., Q.~Zhou, Y.~Shen, et~al.
\newblock Visual chain-of-thought prompting for knowledge-based visual
  reasoning.
\newblock In \emph{AAAI}, pages 1254--1262. 2024.

\bibitem{zhao2023mmicl}
Zhao, H., Z.~Cai, S.~Si, et~al.
\newblock Mmicl: Empowering vision-language model with multi-modal in-context
  learning ability.
\newblock In \emph{ICLR}. 2024.

\bibitem{zheng2023ddcot}
Zheng, G., B.~Yang, J.~Tang, et~al.
\newblock Ddcot: Duty-distinct chain-of-thought prompting for multimodal
  reasoning in language models.
\newblock In \emph{NeurIPS}, pages 5168--5191. 2023.

\bibitem{lu2023chameleon}
Lu, P., B.~Peng, H.~Cheng, et~al.
\newblock Chameleon: Plug-and-play compositional reasoning with large language
  models.
\newblock In \emph{NeurIPS}, pages 43447--43478. 2023.

\bibitem{yao2023react}
Yao, S., J.~Zhao, D.~Yu, et~al.
\newblock React: Synergizing reasoning and acting in language models.
\newblock In \emph{ICLR}. 2024.

\bibitem{tang2023medagents}
Tang, X., A.~Zou, Z.~Zhang, et~al.
\newblock Medagents: Large language models as collaborators for zero-shot
  medical reasoning.
\newblock In \emph{ACL Findings}. 2024.

\bibitem{yang2023mm}
Yang, Z., L.~Li, J.~Wang, et~al.
\newblock Mm-react: Prompting chatgpt for multimodal reasoning and action.
\newblock \emph{arXiv preprint arXiv:2303.11381}, 2023.

\bibitem{qin2023toolllm}
Qin, Y., S.~Liang, Y.~Ye, et~al.
\newblock Toolllm: Facilitating large language models to master 16000+
  real-world apis.
\newblock \emph{arXiv preprint arXiv:2307.16789}, 2023.

\bibitem{liu2023towards}
Liu, X., R.~Li, W.~Ji, et~al.
\newblock Towards robust multi-modal reasoning via model selection.
\newblock In \emph{ICLR}. 2024.

\bibitem{guo2025deepseek}
Guo, D., D.~Yang, H.~Zhang, et~al.
\newblock Deepseek-r1: Incentivizing reasoning capability in llms via
  reinforcement learning.
\newblock \emph{arXiv preprint arXiv:2501.12948}, 2025.

\bibitem{zhang2025srpo}
Zhang, X., J.~Wang, Z.~Cheng, et~al.
\newblock Srpo: A cross-domain implementation of large-scale reinforcement
  learning on llm.
\newblock \emph{arXiv preprint arXiv:2504.14286}, 2025.

\bibitem{wei2025skywork}
Wei, Y., Y.~Peng, X.~Wang, et~al.
\newblock Skywork r1v2: Multimodal hybrid reinforcement learning for reasoning.
\newblock \emph{arXiv preprint arXiv:2504.16656}, 2025.

\bibitem{wang2025visualprm}
Wang, W., Z.~Gao, L.~Chen, et~al.
\newblock Visualprm: An effective process reward model for multimodal
  reasoning.
\newblock \emph{arXiv preprint arXiv:2503.10291}, 2025.

\bibitem{xing2025echoink}
Xing, Z., X.~Hu, C.-W. Fu, et~al.
\newblock Echoink-r1: Exploring audio-visual reasoning in multimodal llms via
  reinforcement learning.
\newblock \emph{arXiv preprint arXiv:2505.04623}, 2025.

\bibitem{yang2025r1}
Yang, Y., X.~He, H.~Pan, et~al.
\newblock R1-onevision: Advancing generalized multimodal reasoning through
  cross-modal formalization.
\newblock \emph{arXiv preprint arXiv:2503.10615}, 2025.

\bibitem{bai2023qwen}
Bai, J., S.~Bai, Y.~Chu, et~al.
\newblock Qwen technical report.
\newblock \emph{arXiv preprint arXiv:2309.16609}, 2023.

\bibitem{grattafiori2024llama}
Grattafiori, A., A.~Dubey, A.~Jauhri, et~al.
\newblock The llama 3 herd of models.
\newblock \emph{arXiv preprint arXiv:2407.21783}, 2024.

\bibitem{team2023gemini}
Team, G., R.~Anil, S.~Borgeaud, et~al.
\newblock Gemini: a family of highly capable multimodal models.
\newblock \emph{arXiv preprint arXiv:2312.11805}, 2023.

\bibitem{wu2023bloomberggpt}
Wu, S., O.~Irsoy, S.~Lu, et~al.
\newblock Bloomberggpt: A large language model for finance.
\newblock \emph{arXiv preprint arXiv:2303.17564}, 2023.

\bibitem{wang2023fingpt}
Wang, N., H.~Yang, C.~D. Wang.
\newblock Fingpt: Instruction tuning benchmark for open-source large language
  models in financial datasets.
\newblock \emph{arXiv preprint arXiv:2310.04793}, 2023.

\bibitem{zhang2023xuanyuan}
Zhang, X., Q.~Yang.
\newblock Xuanyuan 2.0: A large chinese financial chat model with hundreds of
  billions parameters.
\newblock In \emph{CIKM}, pages 4435--4439. 2023.

\bibitem{li2023cfgpt}
Li, J., Y.~Bian, G.~Wang, et~al.
\newblock Cfgpt: Chinese financial assistant with large language model.
\newblock \emph{arXiv preprint arXiv:2309.10654}, 2023.

\bibitem{chen2023disc}
Chen, W., Q.~Wang, Z.~Long, et~al.
\newblock Disc-finllm: A chinese financial large language model based on
  multiple experts fine-tuning.
\newblock \emph{arXiv preprint arXiv:2310.15205}, 2023.

\bibitem{xie2023pixiu}
Xie, Q., W.~Han, X.~Zhang, et~al.
\newblock Pixiu: A large language model, instruction data and evaluation
  benchmark for finance.
\newblock In \emph{NeurIPS}. 2023.

\bibitem{liu2025fin}
Liu, Z., X.~Guo, F.~Lou, et~al.
\newblock Fin-r1: A large language model for financial reasoning through
  reinforcement learning.
\newblock \emph{arXiv preprint arXiv:2503.16252}, 2025.

\bibitem{achiam2023gpt}
Achiam, J., S.~Adler, S.~Agarwal, et~al.
\newblock Gpt-4 technical report.
\newblock \emph{arXiv preprint arXiv:2303.08774}, 2023.

\bibitem{shah2022flue}
Shah, R.~S., K.~Chawla, D.~Eidnani, et~al.
\newblock When flue meets flang: Benchmarks and large pre-trained language
  model for financial domain.
\newblock In \emph{EMNLP}. 2022.

\bibitem{lu2023bbt}
Lu, D., H.~Wu, J.~Liang, et~al.
\newblock Bbt-fin: Comprehensive construction of chinese financial domain
  pre-trained language model, corpus and benchmark.
\newblock \emph{arXiv preprint arXiv:2302.09432}, 2023.

\bibitem{zhang2023fineval}
Zhang, L., W.~Cai, Z.~Liu, et~al.
\newblock Fineval: A chinese financial domain knowledge evaluation benchmark
  for large language models.
\newblock \emph{arXiv preprint arXiv:2308.09975}, 2023.

\bibitem{li2024investorbench}
Li, H., Y.~Cao, Y.~Yu, et~al.
\newblock Investorbench: A benchmark for financial decision-making tasks with
  llm-based agent.
\newblock \emph{arXiv preprint arXiv:2412.18174}, 2024.

\bibitem{chen2021finqa}
Chen, Z., W.~Chen, C.~Smiley, et~al.
\newblock Finqa: A dataset of numerical reasoning over financial data.
\newblock In \emph{EMNLP}. 2021.

\bibitem{xie2024finben}
Xie, Q., W.~Han, Z.~Chen, et~al.
\newblock Finben: A holistic financial benchmark for large language models.
\newblock In \emph{NeurIPS}, pages 95716--95743. 2024.

\bibitem{xu2024superclue}
Xu, L., L.~Zhu, Y.~Wu, et~al.
\newblock Superclue-fin: Graded fine-grained analysis of chinese llms on
  diverse financial tasks and applications.
\newblock \emph{arXiv preprint arXiv:2404.19063}, 2024.

\bibitem{openfindata}
{OpenCompass Project}.
\newblock Openfindata: The open-source financial evaluation dataset for large
  language models, 2023.
\newblock Available at \url{https://github.com/open-compass/OpenFinData}.

\bibitem{xue2025famma}
Xue, S., T.~Chen, F.~Zhou, et~al.
\newblock {FAMMA}: A benchmark for financial multilingual multimodal question
  answering, 2025.

\bibitem{gan2024mme}
Gan, Z., Y.~Lu, D.~Zhang, et~al.
\newblock Mme-finance: A multimodal finance benchmark for expert-level
  understanding and reasoning.
\newblock \emph{arXiv preprint arXiv:2411.03314}, 2024.

\bibitem{yizhao_findataset}
{CMB AILab}.
\newblock Yizhao-findataset, 2024.
\newblock Available at
  \url{https://www.modelscope.cn/datasets/CMB_AILab/YiZhao-FinDataSet}.

\bibitem{liu2024deepseek}
Liu, A., B.~Feng, B.~Xue, et~al.
\newblock Deepseek-v3 technical report.
\newblock \emph{arXiv preprint arXiv:2412.19437}, 2024.

\bibitem{bai2025qwen2}
Bai, S., K.~Chen, X.~Liu, et~al.
\newblock Qwen2.5-vl technical report.
\newblock \emph{arXiv preprint arXiv:2502.13923}, 2025.

\bibitem{shao2024deepseekmath}
Shao, Z., P.~Wang, Q.~Zhu, et~al.
\newblock Deepseekmath: Pushing the limits of mathematical reasoning in open
  language models.
\newblock \emph{arXiv preprint arXiv:2402.03300}, 2024.

\bibitem{yang2025towards}
Yang, W., S.~Ma, Y.~Lin, et~al.
\newblock Towards thinking-optimal scaling of test-time compute for llm
  reasoning.
\newblock \emph{arXiv preprint arXiv:2502.18080}, 2025.

\bibitem{goodfellow2014generative}
Goodfellow, I.~J., J.~Pouget-Abadie, M.~Mirza, et~al.
\newblock Generative adversarial nets.
\newblock In \emph{NeurIPS}. 2014.

\bibitem{luo2025deepscaler}
Luo, M., S.~Tan, J.~Wong, et~al.
\newblock Deepscaler: Surpassing o1-preview with a 1.5 b model by scaling rl.
\newblock \emph{Notion Blog}, 2025.

\bibitem{wang2024measuring}
Wang, K., J.~Pan, W.~Shi, et~al.
\newblock Measuring multimodal mathematical reasoning with math-vision dataset.
\newblock In \emph{NeurIPS}, pages 95095--95169. 2024.

\bibitem{zhang2024mathverse}
Zhang, R., D.~Jiang, Y.~Zhang, et~al.
\newblock Mathverse: Does your multi-modal llm truly see the diagrams in visual
  math problems?
\newblock In \emph{ECCV}, pages 169--186. 2024.

\bibitem{chen2024we}
Chen, L., J.~Li, X.~Dong, et~al.
\newblock Are we on the right way for evaluating large vision-language models?
\newblock In \emph{NeurIPS}. 2024.

\bibitem{lu2023mathvista}
Lu, P., H.~Bansal, T.~Xia, et~al.
\newblock Mathvista: Evaluating mathematical reasoning of foundation models in
  visual contexts.
\newblock In \emph{ICLR}. 2024.

\bibitem{rein2024gpqa}
Rein, D., B.~L. Hou, A.~C. Stickland, et~al.
\newblock Gpqa: A graduate-level google-proof q\&a benchmark.
\newblock In \emph{COLM}. 2024.

\bibitem{xue2025fammabenchmarkfinancialdomain}
Xue, S., X.~Li, F.~Zhou, et~al.
\newblock Famma: A benchmark for financial domain multilingual multimodal
  question answering.
\newblock \emph{arXiv preprint arXiv:2410.04526}, 2025.

\bibitem{chen2024expanding}
Chen, Z., W.~Wang, Y.~Cao, et~al.
\newblock Expanding performance boundaries of open-source multimodal models
  with model, data, and test-time scaling.
\newblock \emph{arXiv preprint arXiv:2412.05271}, 2024.

\bibitem{yao2024minicpm}
Yao, Y., T.~Yu, A.~Zhang, et~al.
\newblock Minicpm-v: A gpt-4v level mllm on your phone.
\newblock \emph{arXiv preprint arXiv:2408.01800}, 2024.

\bibitem{lu2021iconqa}
Lu, P., L.~Qiu, J.~Chen, et~al.
\newblock Iconqa: A new benchmark for abstract diagram understanding and visual
  language reasoning.
\newblock \emph{arXiv preprint arXiv:2110.13214}, 2021.

\bibitem{kembhavi2017you}
Kembhavi, A., M.~Seo, D.~Schwenk, et~al.
\newblock Are you smarter than a sixth grader? textbook question answering for
  multimodal machine comprehension.
\newblock In \emph{CVPR}, pages 4999--5007. 2017.

\bibitem{zhang2023pmc}
Zhang, X., C.~Wu, Z.~Zhao, et~al.
\newblock Pmc-vqa: Visual instruction tuning for medical visual question
  answering.
\newblock \emph{arXiv preprint arXiv:2305.10415}, 2023.

\bibitem{mathew2022infographicvqa}
Mathew, M., V.~Bagal, R.~Tito, et~al.
\newblock Infographicvqa.
\newblock In \emph{WACV}, pages 1697--1706. 2022.

\bibitem{lu2022dynamic}
Lu, P., L.~Qiu, K.-W. Chang, et~al.
\newblock Dynamic prompt learning via policy gradient for semi-structured
  mathematical reasoning.
\newblock \emph{arXiv preprint arXiv:2209.14610}, 2022.

\bibitem{singh2019towards}
Singh, A., V.~Natarajan, M.~Shah, et~al.
\newblock Towards vqa models that can read.
\newblock In \emph{CVPR}, pages 8317--8326. 2019.

\bibitem{kembhavi2016diagram}
Kembhavi, A., M.~Salvato, E.~Kolve, et~al.
\newblock A diagram is worth a dozen images.
\newblock In \emph{ECCV}, pages 235--251. 2016.

\bibitem{goyal2017making}
Goyal, Y., T.~Khot, D.~Summers-Stay, et~al.
\newblock Making the v in vqa matter: Elevating the role of image understanding
  in visual question answering.
\newblock In \emph{CVPR}, pages 6904--6913. 2017.

\bibitem{kahou2017figureqa}
Kahou, S.~E., V.~Michalski, A.~Atkinson, et~al.
\newblock Figureqa: An annotated figure dataset for visual reasoning.
\newblock In \emph{ICLR workshop}. 2018.

\bibitem{masry2022chartqa}
Masry, A., D.~X. Long, J.~Q. Tan, et~al.
\newblock Chartqa: A benchmark for question answering about charts with visual
  and logical reasoning.
\newblock In \emph{ACL}. 2022.

\bibitem{lu2022learn}
Lu, P., S.~Mishra, T.~Xia, et~al.
\newblock Learn to explain: Multimodal reasoning via thought chains for science
  question answering.
\newblock In \emph{NeurIPS}, pages 2507--2521. 2022.

\bibitem{methani2020plotqa}
Methani, N., P.~Ganguly, M.~M. Khapra, et~al.
\newblock Plotqa: Reasoning over scientific plots.
\newblock In \emph{WACV}, pages 1527--1536. 2020.

\bibitem{cao2022augmented}
Cao, J., J.~Xiao.
\newblock An augmented benchmark dataset for geometric question answering
  through dual parallel text encoding.
\newblock In \emph{COLING}, pages 1511--1520. 2022.

\bibitem{li2023super}
Li, Z., X.~Wang, E.~Stengel-Eskin, et~al.
\newblock Super-clevr: A virtual benchmark to diagnose domain robustness in
  visual reasoning.
\newblock In \emph{CVPR}, pages 14963--14973. 2023.

\bibitem{kafle2018dvqa}
Kafle, K., B.~Price, S.~Cohen, et~al.
\newblock Dvqa: Understanding data visualizations via question answering.
\newblock In \emph{CVPR}, pages 5648--5656. 2018.

\bibitem{antol2015vqa}
Antol, S., A.~Agrawal, J.~Lu, et~al.
\newblock Vqa: Visual question answering.
\newblock In \emph{ICCV}, pages 2425--2433. 2015.

\bibitem{lu2021inter}
Lu, P., R.~Gong, S.~Jiang, et~al.
\newblock Inter-gps: Interpretable geometry problem solving with formal
  language and symbolic reasoning.
\newblock In \emph{ACL}. 2021.

\bibitem{chang2022mapqa}
Chang, S., D.~Palzer, J.~Li, et~al.
\newblock Mapqa: A dataset for question answering on choropleth maps.
\newblock In \emph{NeurIPS}. 2022.

\bibitem{chen2022unigeo}
Chen, J., T.~Li, J.~Qin, et~al.
\newblock Unigeo: Unifying geometry logical reasoning via reformulating
  mathematical expression.
\newblock In \emph{EMNLP}. 2022.

\bibitem{gurari2018vizwiz}
Gurari, D., Q.~Li, A.~J. Stangl, et~al.
\newblock Vizwiz grand challenge: Answering visual questions from blind people.
\newblock In \emph{CVPR}, pages 3608--3617. 2018.

\bibitem{seo2015solving}
Seo, M., H.~Hajishirzi, A.~Farhadi, et~al.
\newblock Solving geometry problems: Combining text and diagram interpretation.
\newblock In \emph{EMNLP}, pages 1466--1476. 2015.

\bibitem{lindstrom2022clevr}
Lindstr{\"o}m, A.~D., S.~S. Abraham.
\newblock Clevr-math: A dataset for compositional language, visual and
  mathematical reasoning.
\newblock \emph{arXiv preprint arXiv:2208.05358}, 2022.

\bibitem{lau2018dataset}
Lau, J.~J., S.~Gayen, A.~Ben~Abacha, et~al.
\newblock A dataset of clinically generated visual questions and answers about
  radiology images.
\newblock \emph{Scientific data}, 5(1):1--10, 2018.

\bibitem{zheng2025easyr1}
Yaowei, Z., L.~Junting, W.~Shenzhi, et~al.
\newblock Easyr1: An efficient, scalable, multi-modality rl training framework.
\newblock \url{https://github.com/hiyouga/EasyR1}, 2025.

\bibitem{sheng2024hybridflow}
Sheng, G., C.~Zhang, Z.~Ye, et~al.
\newblock Hybridflow: A flexible and efficient rlhf framework.
\newblock \emph{arXiv preprint arXiv: 2409.19256}, 2024.

\bibitem{reimers2019sentence}
Reimers, N., I.~Gurevych.
\newblock Sentence-bert: Sentence embeddings using siamese bert-networks.
\newblock In \emph{EMNLP-IJCNLP}, pages 3982--3992. 2019.

\bibitem{devlin2019bert}
Devlin, J., M.-W. Chang, K.~Lee, et~al.
\newblock Bert: Pre-training of deep bidirectional transformers for language
  understanding.
\newblock In \emph{NAACL}, pages 4171--4186. 2019.

\end{thebibliography}
}

\newpage

\appendix

The appendix includes additional materials to supplement the main submission.
Section~\ref{sec:training} details the training data and implementation settings.
Section~\ref{sec:evaluation} describes the evaluation benchmarks and metrics.
Section~\ref{sec:more_exp} presents additional experimental results on the FinLMM-R1 framework and training process.
Section~\ref{sec:statistics} provides further statistics on the ASP-derived FinData.
Section~\ref{sec:prompt} lists the prompts used during training and evaluation.
Section~\ref{sec:human_instruction} outlines the instructions given to human annotators for evaluating the quality of ASP-derived FinData.
Section~\ref{sec:case} presents additional case studies illustrating the effectiveness of FinLMM-R1.
Section~\ref{sec:stat} reports significance tests comparing our method with the strongest baselines.
Section~\ref{sec:lim} discusses the limitations and future directions of our work.
Section~\ref{sec:imp} discusses the broader impacts of our work.

\section{Training Details~\label{sec:training}}
\subsection{Datasets}

In the first training stage, we use the \textit{DeepScaler-Preview Dataset} as the reasoning training set, which is a text-only dataset containing 40,300 unique math problems along with their corresponding solutions.
In the second stage, we further train the model on the \textit{VerMulti-65K} dataset provided by LMM-R1, which comprises problems collected from 24 different sources. 
The data distribution is shown in Table~\ref{tab:dataset_counts}.
In addition, we select 40,000 samples from our ASP-derived financial multimodal dataset (\emph{i.e.}, FinData) and combine them with \textit{VerMulti-65K} for training, aiming to validate the effectiveness and scalability of our dataset in financial multimodal reasoning.
More details about the dataset statistics are provided in Sec.~\ref{sec:statistics}.

\begin{table}[htbp]
\centering
\caption{Dataset composition and instance counts for each split}
\label{tab:dataset_counts}
\begin{tabular}{ll|ll}
\toprule
\textbf{Split} & \textbf{Count} & \textbf{Split} & \textbf{Count} \\
\midrule
IconQA~\cite{lu2021iconqa} & 7166 & TQA~\cite{kembhavi2017you} & 2130 \\
PMC-VQA~\cite{zhang2023pmc} & 6760 & DocVQA~\cite{mathew2022infographicvqa} & 1974 \\
TabMWP~\cite{lu2022dynamic} & 6732 & TextVQA~\cite{singh2019towards} & 1462 \\
A-OKVQA~\cite{kembhavi2016diagram} & 6185 & VQA2.0~\cite{goyal2017making} & 1316 \\
FigureQA~\cite{kahou2017figureqa} & 4995 & ChartQA~\cite{masry2022chartqa} & 1115 \\
ScienceQA~\cite{lu2022learn} & 4243 & PlotQA~\cite{methani2020plotqa} & 1020 \\
GeoQA+~\cite{cao2022augmented} & 4062 & Super-CLEVR~\cite{li2023super} & 1016 \\
DVQA~\cite{kafle2018dvqa} & 3317 & VQA-AS~\cite{antol2015vqa} & 404 \\
Geometry3K~\cite{lu2021inter} & 2845 & MapQA~\cite{chang2022mapqa} & 278 \\
UniGeo~\cite{chen2022unigeo} & 2767 & VizWiz~\cite{gurari2018vizwiz} & 204 \\
AI2D~\cite{kembhavi2016diagram} & 2603 & GEOS~\cite{seo2015solving} & 129 \\
CLEVR-Math~\cite{lindstrom2022clevr} & 2393 & VQA-RAD~\cite{lau2018dataset} & 2 \\
\bottomrule
\end{tabular}
\end{table}

\subsection{Implementation}
We use the lightweight and efficient \textbf{Easy-R1}\cite{zheng2025easyr1} framework, implemented by Verl~\cite{sheng2024hybridflow}, for training and inference.
It supports multiple reinforcement learning algorithms and is well suited for multi-stage training and multimodal tasks.
All experiments are conducted on eight NVIDIA A100 GPUs with 80GB memory each~(\emph{i.e.}, NVIDIA DGX A100 640G).
The \textbf{GRPO algorithm} is employed in both training stages.
All models in the main submission are initialized from \textbf{Qwen2.5-VL-Instruct-3B}.
The main hyperparameter settings for each stage are listed in Table~\ref{tab:finlmm_hyperparams}.

\begin{table}[htbp]
\centering
\caption{Main hyperparameter settings for each stage}
\label{tab:finlmm_hyperparams}
\resizebox{\textwidth}{!}{
\setlength\tabcolsep{3pt}
\begin{tabular}{lcccc}
\toprule
\textbf{Parameter} & \textbf{FinLMM-R1 (S1)} & \textbf{FinLMM-R1 (S2)} & \textbf{FinLMM-R1 (S2) + FinData} \\
\midrule
train\_batch\_size & 160 & 160 & 160 \\
rollout\_batch\_size & 320 & 320 & 320 \\
num\_epochs & 1 & 1 & 1 \\
n\_samples\_per\_prompt & 16 & 16 & 16 \\
max\_response\_length & 2048 & 2048 & 2048 \\
max\_prompt\_length & 2048 & 10240 & 10240 \\
optimizer & adamw\_bf16 & adamw\_bf16 & adamw\_bf16 \\
temperature & 1 & 1 & 1 \\
kl\_coef & 1e-2 & 1e-2 & 1e-2 \\
actor\_learning\_rate & 1e-6 & 1e-6 & 1e-6 \\
warm-up ratio & 0 & 0 & 0 \\
training\_time (h) & 12 & 86 & 140 \\
format\_weight & 0.5 & 0.1 & 0.1 \\
accuracy\_weight & 0.5 & 0.5 & 0.5 \\
bert\_reward\_weight & / & 0.15 & 0.15 \\
image\_selection\_weight & / & 0.2 & 0.2 \\
length\_weight & / & 0.05 & 0.05 \\
bert\_model & / & sentence\_BERT base & sentence\_BERT base \\
bert\_training\_epochs & / & 1 & 1 \\
bert\_batch\_size & / & 128 & 128 \\
bert\_learning\_rate & / & 5e-6 & 5e-6 \\
bert\_train\_interval (steps) & / & 10 & 10 \\
minimum\_target\_lengths (token) & / & 300 & 300 \\
optimal\_target\_lengths (token) & / & 450 & 450 \\
maximum\_target\_lengths (token) & / & 600 & 600 \\
\bottomrule
\end{tabular}
}
\end{table}

\section{Evaluation Details~\label{sec:evaluation}}

\subsection{Text-Only Benchmarks}

For text-only evaluation, we use \textit{GPQA-main}~\cite{rein2024gpqa} to assess the reasoning ability of FinLMM-R1 in purely textual contexts.
The dataset contains multiple-choice questions in biology, physics, and chemistry, authored by domain experts, with accuracy reported in our experiments.

\subsection{General Multimodal Benchmarks}

We evaluate our model on four multimodal reasoning datasets, including \textit{MathVision}-test~\cite{wang2024measuring}, \textit{MathVerse}-testmini~\cite{zhang2024mathverse}, \textit{MM-Star}~\cite{chen2024we}, and \textit{MathVista}-testmini~\cite{lu2023mathvista}, to assess the reasoning ability of FinLMM-R1 in general-purpose multimodal scenarios.
\textit{MathVision} consists of math problems with visual contexts, all sourced from real mathematics competitions and covering 16 distinct mathematical domains.
\textit{MathVerse} contains high-quality problems spanning a wide range of mathematical subjects.
\textit{MM-Star} includes test samples, each designed with strong visual dependency and minimal risk of data leakage.
\textit{MathVista} contains samples across three categories: IQTest, FunctionQA, and PaperQA, designed to evaluate logical reasoning in diagram-based puzzles, algebraic reasoning over function plots, and scientific reasoning based on academic figures, respectively.
All evaluation datasets, including both text-only and multimodal formats, are evaluated using accuracy as the primary metric.
To accommodate the diversity of answer formats across different question types, we design a targeted scoring mechanism.
For multiple-choice questions, the system supports flexible answer recognition, including single-letter options (A/B/C/D), full-format responses (\emph{e.g.}, "A. content"), or the option content alone.
For open-ended questions, we apply normalization techniques that unify mathematical symbol representations, handle variations in fractions and units, and ignore irrelevant formatting, to determine equivalence between model predictions and reference answers and ensure fair evaluation.

\subsection{Financial Multimodal Benchmarks}

For evaluating financial multimodal reasoning, we construct a test set based on the ASP-derived FinData from 2023.
Three graduate students with a background in finance are invited to annotate the samples in terms of image–question alignment and question-answer answerability.
As a result, we obtain 3,823 IQA triplets covering four categories: \textit{Arithmetic Reasoning}~\textbf{(AR)}, \textit{Statistical Reasoning}~\textbf{(SR)}, \textit{Financial Explanation}~\textbf{(FE)}, and \textit{Financial Knowledge}~\textbf{(FK)}.
In addition, we utilize the \textit{FAMMA}-basic~\cite{xue2025fammabenchmarkfinancialdomain} dataset and select multimodal samples with multiple-choice and open-ended questions, as an additional evaluation subset.

\section{More Experimental Results~\label{sec:more_exp}}
\subsection{Effect of Discriminator BERT}

\begin{table}[t]
\caption{Ablation study on different discriminator BERT.}
\label{tab:exp:abl}
\centering
\setlength\tabcolsep{6pt}
\resizebox{\textwidth}{!}{
\begin{tabular}{l|cccc|cc}
\toprule
\multirow{2}{*}{Model} & \multicolumn{4}{c|}{General}                 & \multicolumn{2}{c}{Financial} \\ \addlinespace[2pt]\cline{2-7} \addlinespace[2pt]
                       & MathVision & MathVerse & MM-Star & MathVista & Ours      & FAMMA         \\ \midrule
LMM-R1                 &  26.35     & 41.55     & 58.00   & 63.20     & 53.01     & 26.59         \\
FinLMM-R1(S1)          &  25.63     & 38.39     & 54.20   & 61.52     & 50.73     & 25.35         \\
FinLMM-R1(S2)+SBERT-B  &  27.07     & 42.29     & 61.13   & 65.80     & 55.02     & 27.49         \\
FinLMM-R1(S2)+SBERT-L  &  27.89     & 42.87     & 61.80   & 66.70     & 56.60     & 27.62         \\
FinLMM-R1(S2)+BERT-B   &  26.61     & 41.73     & 60.00   & 65.30     & 55.09     & 27.13        \\
FinLMM-R1(S2)+BERT-L   &  26.95     & 42.11     & 60.20   & 66.30     & 55.33     & 27.96              \\ \bottomrule
\end{tabular}
}
\end{table}

In Table~\ref{tab:exp:abl}, we further investigate the impact of different discriminator BERT variants.
In the main submission, we employ Sentence-BERT-base~(SBERT)~\cite{reimers2019sentence} as the discriminator for all experiments.
In this section, we analyze the effects of both the BERT variant~(SBERT~\cite{reimers2019sentence} \emph{v.s.} BERT~\cite{devlin2019bert}) and the model size~(base \emph{v.s.} large).
When comparing model sizes, we observe that SBERT-large consistently outperforms SBERT-base across all tasks, with gains ranging from 0.5\% to 1.5\%. This suggests that larger discriminators offer better capability in reasoning trace assessment, thereby contributing to improved performance of FinLMM-R1.
When comparing BERT variants, we find that SBERT outperforms BERT on most tasks, as SBERT is specifically optimized for sentence-level tasks, resulting in stronger sentence representation and better support for reasoning LMM training.

\subsection{Experiments on QwenVL-2.5-7B}

\begin{table}[h]
\caption{Experimental results on general multimodal reasoning tasks initialized from Qwen2.5-VL-7B. Best results are in \textbf{bolded}.}
\label{tab:exp:gmm}
\centering
\setlength\tabcolsep{7pt}
\resizebox{\textwidth}{!}{
\begin{tabular}{l|c|cccc|c}
\toprule
\multirow{2}{*}{Model}            & \multicolumn{1}{c|}{Text-only} & \multicolumn{4}{c|}{Multimodal}  & \multirow{2}{*}{Average} \\ \addlinespace[2pt]\cline{2-6} \addlinespace[2pt]
                      & GPQA         & MathVision   & MathVerse    & MM-Star      & MathVista    &        \\ \midrule
Qwen2.5-VL-7B CoT     & 30.91        & 25.10        & 37.36        & 61.07        & 68.20        & 44.53  \\
LMM-R1                & 39.27        & 29.05        & 42.41        & 64.33        & 68.90        & 48.79  \\
FinLMM-R1(S1)         & 37.46        & 27.39        & 40.17        & 63.27        & 67.50        & 47.16  \\
FinLMM-R1(S2)         &\textbf{44.31}& 30.53        & 43.61        &\textbf{66.47}& 71.20        & \textbf{51.22}  \\
FinLMM-R1(S2)+FinData & 43.64        &\textbf{30.68}&\textbf{43.93}& 65.67        &\textbf{71.80}& 51.14  \\ \bottomrule
\end{tabular}
}
\end{table}

\begin{table}[h]
\caption{Experimental results on financial multimodal reasoning tasks initialized from Qwen2.5-VL-7B. Best results are in \textbf{bolded}.}
\label{tab:exp:fmm}
\centering
\setlength\tabcolsep{6pt}
\resizebox{\textwidth}{!}{
\begin{tabular}{l|ccccc|cccc}
\toprule
\multirow{2}{*}{Model} & \multicolumn{5}{c|}{Ours}              & \multicolumn{4}{c}{FAMMA}     \\ \addlinespace[2pt]\cline{2-10} \addlinespace[2pt]
                       & AR    & SR    & FE    & FK    & Tot.  & Easy  & Middle & Hard  & Tot.  \\ \midrule
Qwen2.5-VL-7B CoT      & 51.01 & 52.11 & 52.78 & 77.02 & 57.34 & 37.50 & 26.75  & 15.34 & 28.56 \\
LMM-R1                 & 56.73 & 56.32 & 56.75 & 79.84 & 61.72 & 39.74 & 28.66  & 19.05 & 31.15 \\
LMM-R1+FinData         & 57.88 & 58.95 & 60.85 & 82.45 & 64.29 & 39.42 & 29.75  & 19.98 & 31.53 \\
FinLMM-R1(S1)          & 48.99 & 48.95 & 55.29 & 79.23 & 57.65 & 35.90 & 21.66  & 17.46 & 27.21 \\
FinLMM-R1(S2)          & 57.33 & 59.21 & 58.99 & 79.22 & 62.80 & 39.29 & 29.33  & 20.34 & 31.47 \\
FinLMM-R1(S2)+FinData  & \textbf{59.78} & \textbf{62.63} & \textbf{61.19} & \textbf{83.27} & \textbf{65.59} & \textbf{40.71} & \textbf{30.57}  & \textbf{21.13} & \textbf{32.67} \\ \bottomrule
\end{tabular}
}
\end{table}

Tables~\ref{tab:exp:gmm} and~\ref{tab:exp:fmm} further evaluate our method using \texttt{Qwen2.5-VL-7B} across both general and financial multimodal reasoning tasks.
Consistent with the main results, our enhanced reward framework and the ASP-derived FinData lead to consistent performance improvements in both domains, with gains ranging from 1\% to 5\%.
Specifically, in the general domain~(Table~\ref{tab:exp:gmm}), FinLMM-R1(S2) achieves over a 3\% gain on text reasoning tasks compared to Stage-1 training, and outperforms LMM-R1 across all four multimodal benchmarks. 
MM-Star and MathVista yield higher gains than MathVision and MathVerse, which may be attributed to the relative difficulty of adversarial reward discrimination in more complex reasoning tasks.
In the financial domain~(Table~\ref{tab:exp:fmm}), FinLMM-R1(S2) surpasses LMM-R1 by over 2\% on ASP-derived datasets, and the inclusion of FinData further improves average performance by an additional 2\%. 
These results validate the effectiveness of our reward framework and ASP data construction in enhancing the reasoning ability of LMMs across both general and domain-specific settings.

\subsection{Training Curves}

\begin{figure}[h]
    \centering
    \begin{subfigure}{0.49\textwidth}
        \includegraphics[width=\linewidth]{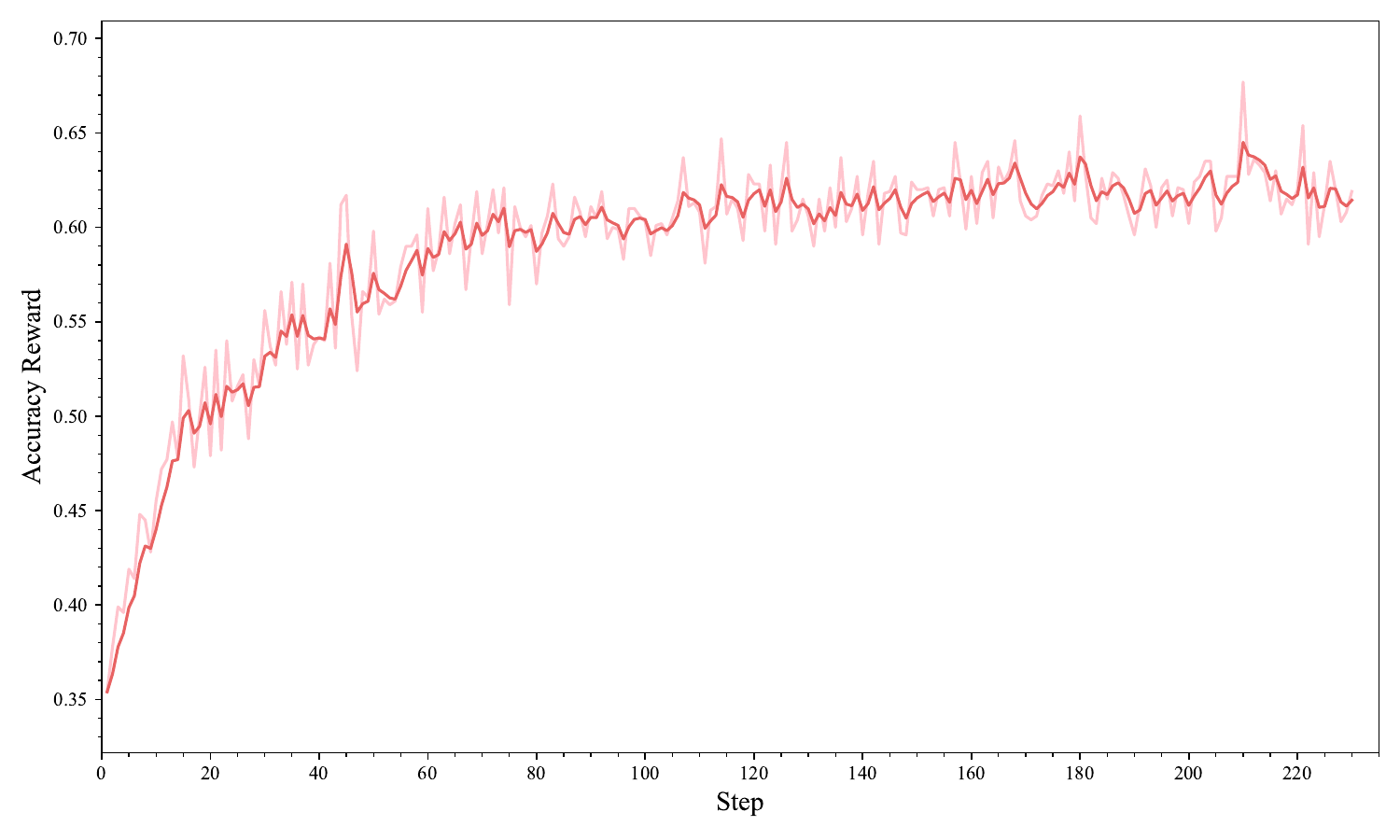}
        \caption{The Accuracy Reward Curve}
    \end{subfigure}
    \hfill
    \begin{subfigure}{0.49\textwidth}
        \includegraphics[width=\linewidth]{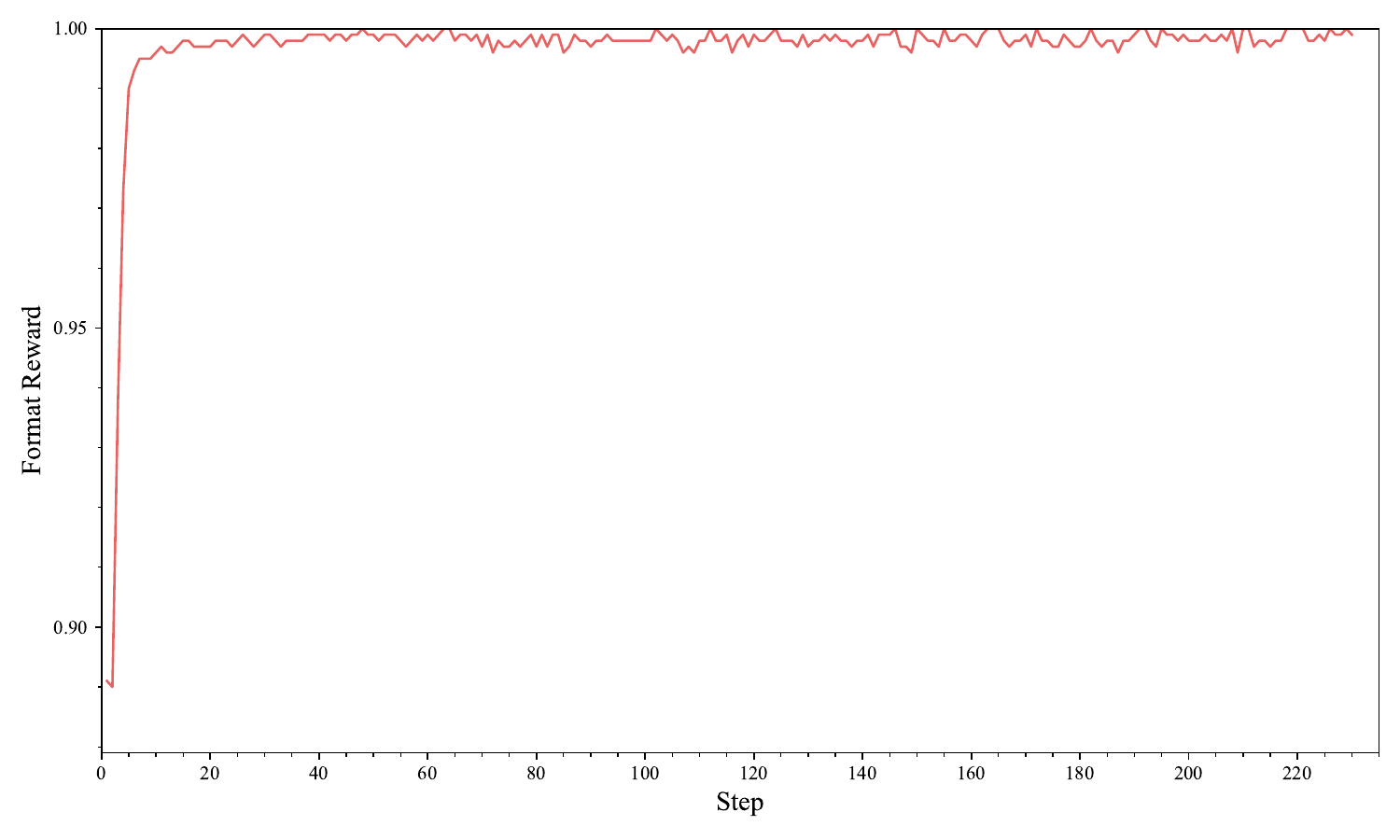}
        \caption{The Format Reward Curve}
    \end{subfigure}

    \vspace{0.5cm} 
    \begin{subfigure}{0.49\textwidth}
        \includegraphics[width=\linewidth]{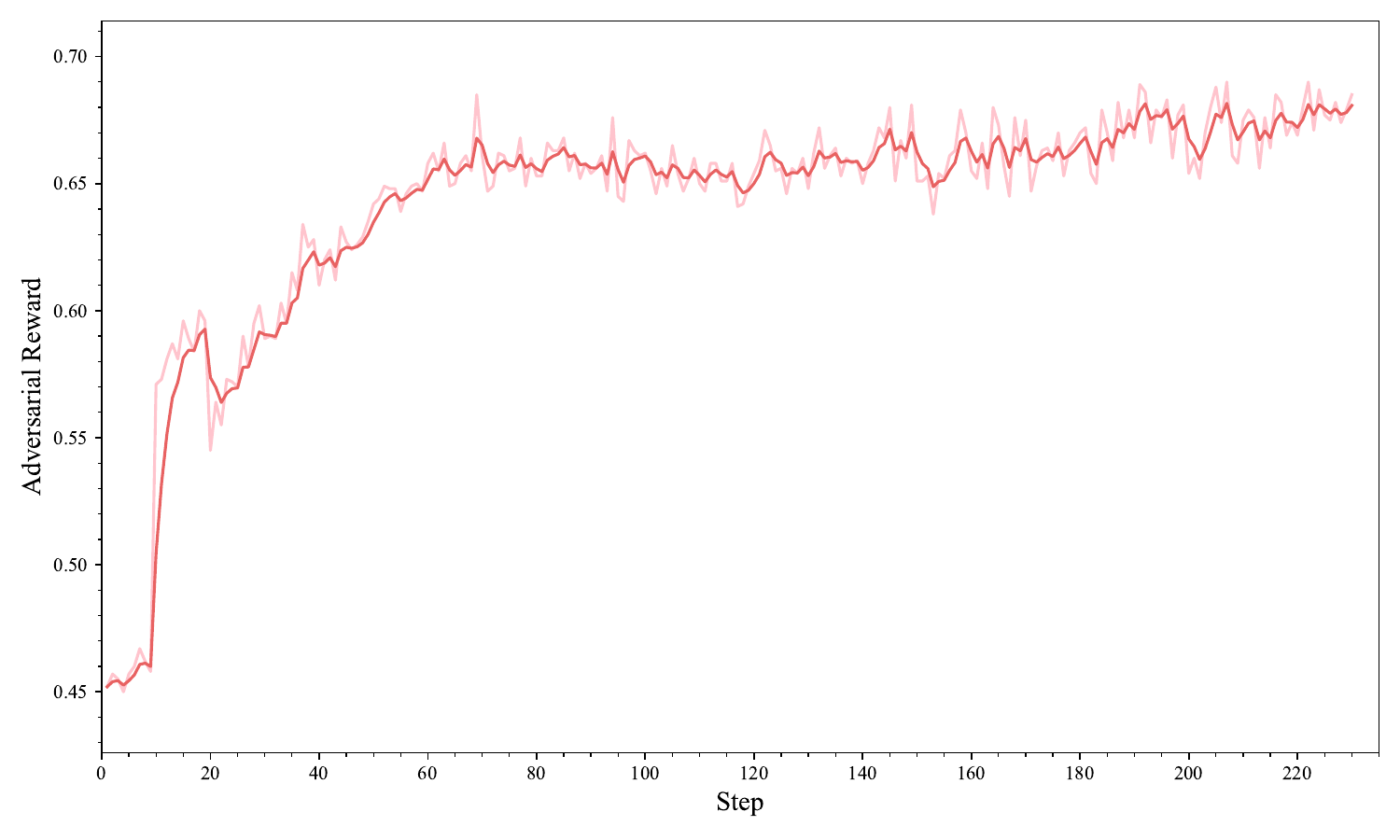}
        \caption{The Adversarial Reward Curve}
    \end{subfigure}
    \hfill
    \begin{subfigure}{0.49\textwidth}
        \includegraphics[width=\linewidth]{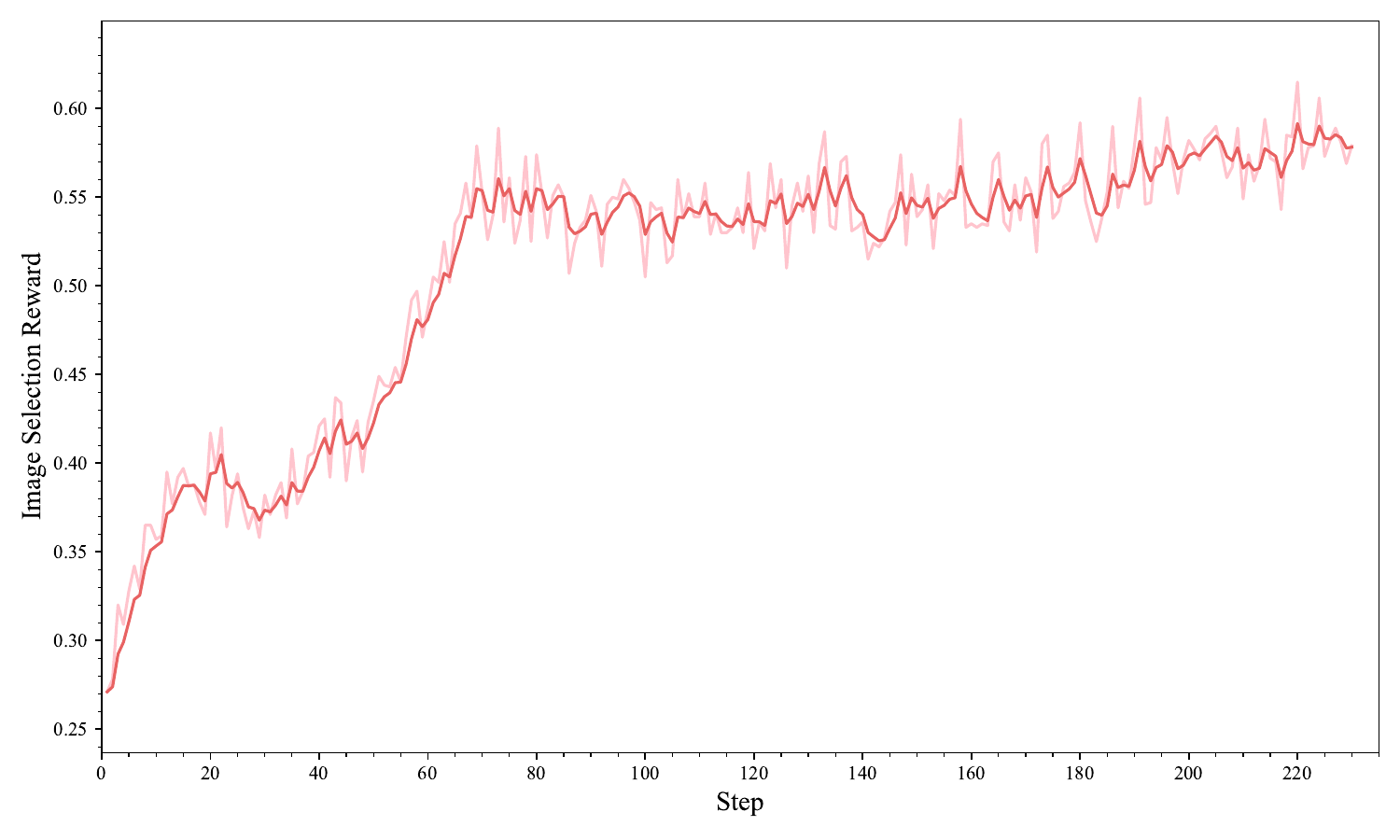}
        \caption{The Image Selection Reward Curve}
    \end{subfigure}

    \vspace{0.5cm} 
    \begin{subfigure}{0.49\textwidth}
        \includegraphics[width=\linewidth]{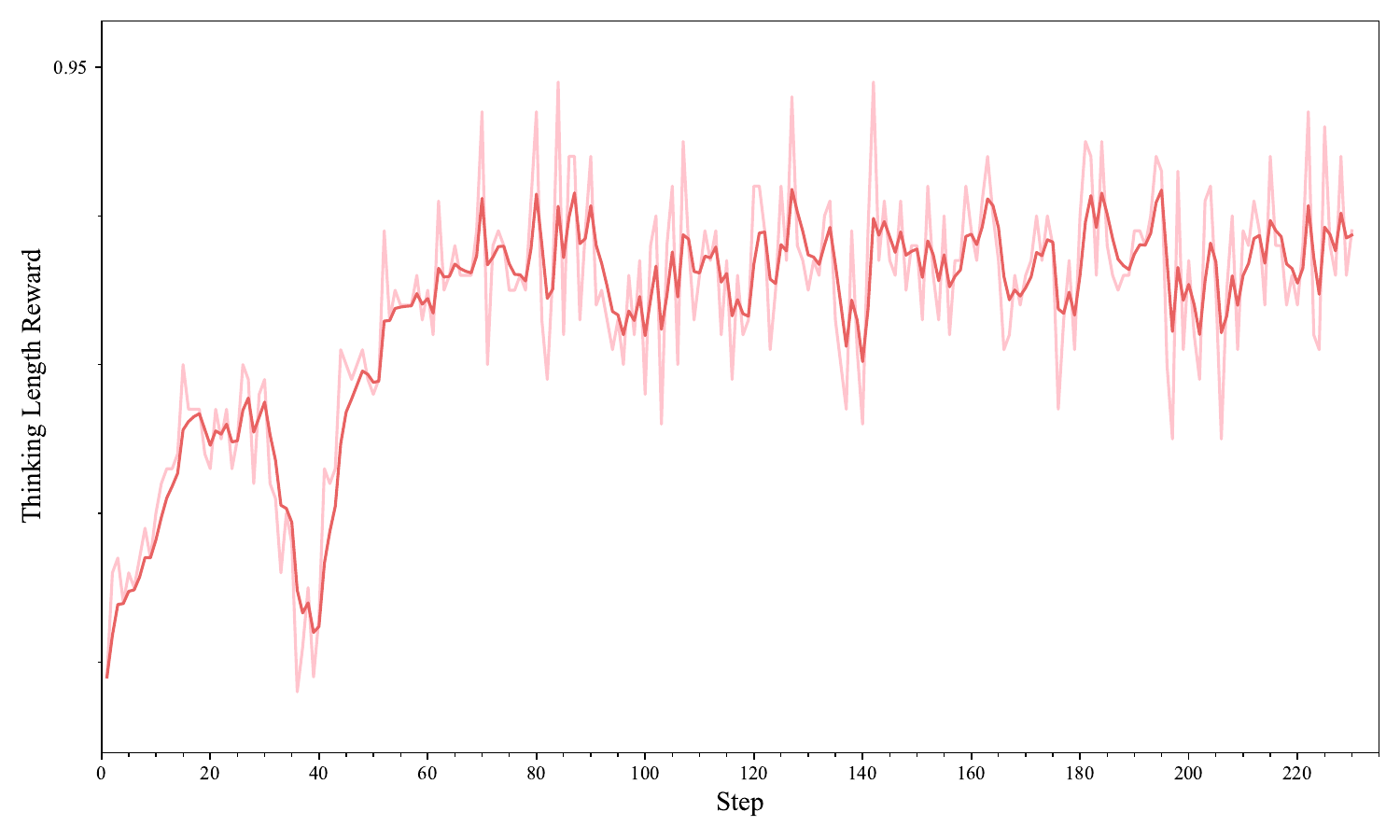}
        \caption{The Length Reward Curve}
    \end{subfigure}
    \hfill
    \begin{subfigure}{0.49\textwidth}
        \includegraphics[width=\linewidth]{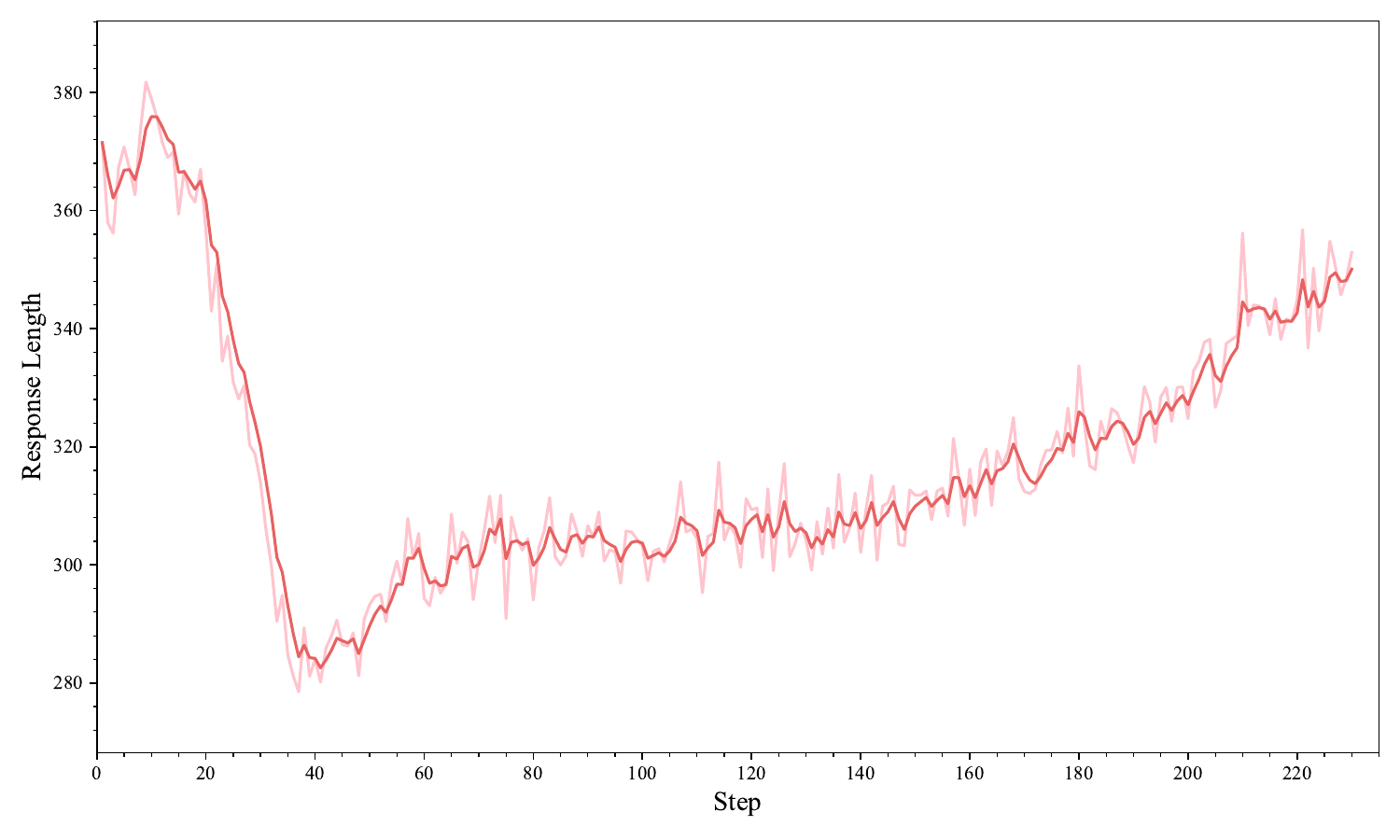}
        \caption{The Response Length Curve}
    \end{subfigure}

    \caption{The Training Curve of FinLMM-R1 in Stage-2.}
    \label{fig:3x2layout}
\end{figure}

In Figure~\ref{fig:3x2layout}, we present the variance of different reward signals and the response length of FinLMM-R1 during the multimodal training phase~(Stage-2).
The accuracy and format rewards gradually increase at the beginning of training and then stabilize, where accuracy reaches around 0.6 and the format reward approaches 1.0, indicating the effectiveness of these two original reward components.
For adversarial, length, and image selection rewards, we observe a short-term drop between steps 20 and 40, which correlates with fluctuations in response length.
As the model transitions from text-only reasoning in Stage-1 to multimodal reasoning in Stage-2, the response length initially decreases due to the combined influence of the accuracy reward and the inherent modality shift.
This reduction in response length temporarily compromises the ability of LMM to generate comprehensive analyses, thereby causing a decline in adversarial, length, and image selection rewards.
However, as training progresses, the demands of different rewards on response length become increasingly aligned.
Despite shorter responses, each reward signal gradually guides the model toward generating outputs that balance informativeness, structural clarity, and visual grounding.
This synergy allows all reward curves to recover and improve steadily after step 40, indicating that the model successfully adapts to the multimodal reasoning setting and learns to optimize multiple objectives simultaneously.

\section{Data Statistics~\label{sec:statistics}}
\begin{figure}[h]
\centering
\includegraphics[width=0.85\textwidth]{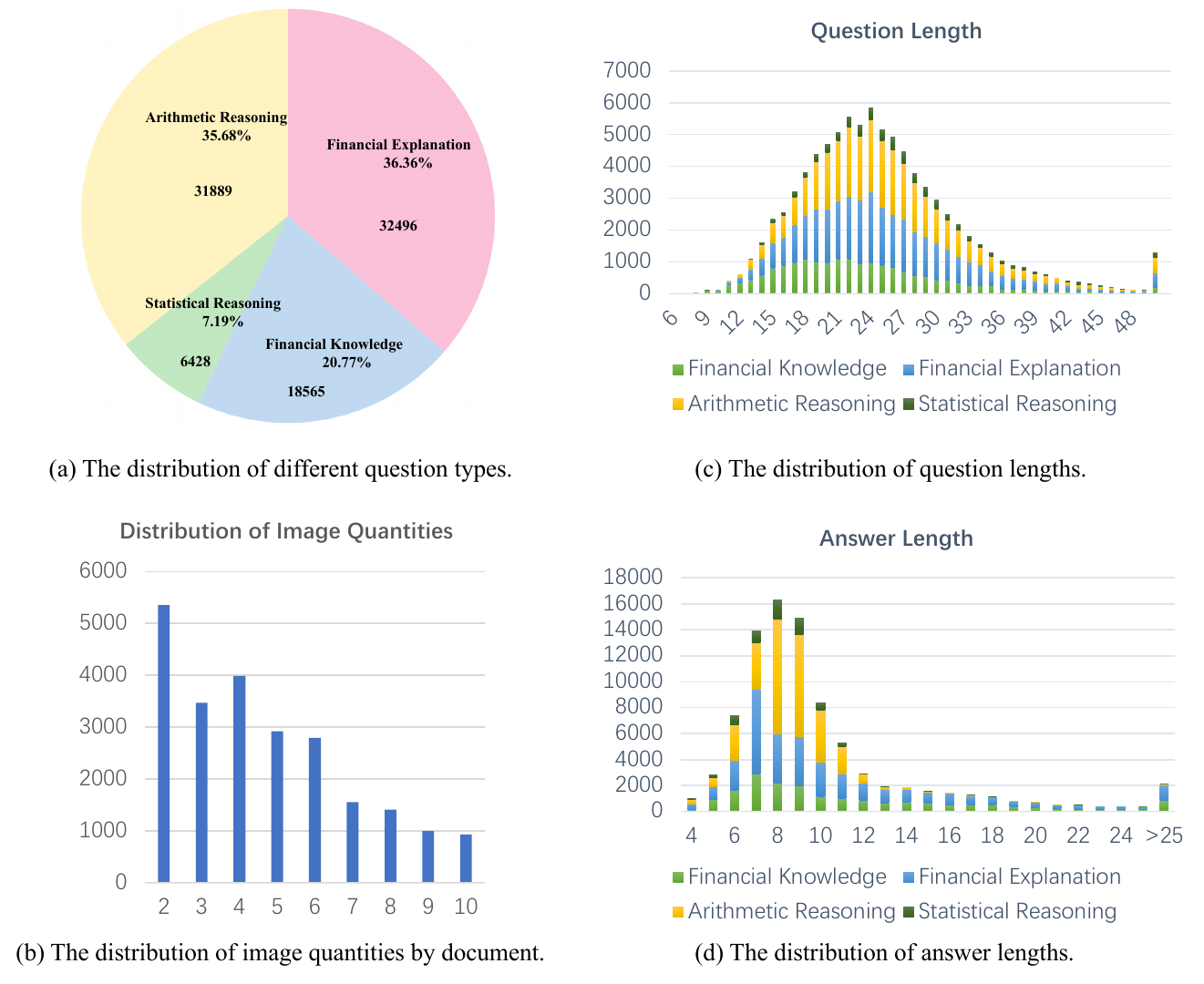} 
\caption{The statistics of FinData in (a) question type, (b) image quantity distribution by document, (c) question length distribution, and (d) answer length distribution. Best viewed by zoom in.}
\label{fig:data-stat}
\end{figure}

Our ASP-derived  FinData is derived from a large collection of financial report PDFs, totaling 23,397 documents.
These reports yield 89,378 questions and 108,738 images.
Detailed statistics are provided in Tables~\ref{tab:data-stat1} and~\ref{tab:data-stat2}, and Figure~\ref{fig:data-stat}.
Table~\ref{tab:data-stat1} presents detailed statistics on the total number of PDFs, questions, images, and the distribution of question types.
On average, each PDF contains approximately 3.82 questions and 4.65 images.
The images span a diverse range of visual forms commonly found in financial reports, such as tables, line charts, bar charts, and pie charts.
Table~\ref{tab:data-stat2} provides further insight into the textual characteristics of the dataset, presenting the average lengths of question descriptions and answer options across four categories: \textit{Arithmetic Reasoning}, \textit{Statistical Reasoning}, \textit{Financial Knowledge}, and \textit{Financial Explanation}.
These statistics offer valuable insights into the complexity and variability of the questions.
Importantly, the dataset is organized into image–question–answer triplets, where each image corresponds to a visual element extracted from a financial report, and each question targets specific information or reasoning based on that image.
This multimodal format facilitates comprehensive financial understanding by integrating both visual and textual information.
Overall, the scale and multimodal diversity of the dataset provide a robust foundation for developing and training large-scale financial multimodal reasoning models.

\begin{table}[ht]
\centering
\caption{Dataset Statistics}
\label{tab:data-stat1}
\begin{tabular}{lc}
\toprule
\textbf{Metric} & \textbf{Value} \\
\midrule
Total number of financial report PDFs     & 23,397 \\
Total number of questions                 & 89,378 \\
Total number of images                    & 108,738 \\
Arithmetic Reasoning questions            & 31,889 \\
Statistical Reasoning questions           & 6,428 \\
Financial Knowledge questions             & 18,565 \\
Financial Explanation questions           & 32,496 \\
Average number of questions per PDF       & 3.82 \\
Average number of images per PDF          & 4.65 \\
\bottomrule
\end{tabular}
\end{table}

\begin{table}[ht]
\centering
\caption{Average Lengths of Questions and Options by Category}
\label{tab:data-stat2}
\begin{tabular}{lcc}
\toprule
\textbf{Category} & \textbf{Average Question Length} & \textbf{Average Option Length} \\
\midrule
Arithmetic Reasoning     & 25.81 & 8.65 \\
Statistical Reasoning    & 27.81 & 8.51 \\
Financial Knowledge      & 23.20 & 11.04 \\
Financial Explanation    & 25.52 & 10.57 \\
\bottomrule
\end{tabular}
\end{table}

\section{Prompts for Training and Evaluation~\label{sec:prompt}}

\begin{tcolorbox}[colback=gray!15, colframe=black, title=Prompts for Text-only Training and Evaluation, fonttitle=\bfseries, breakable]

A conversation between the User and the Assistant. The User asks a question, and the Assistant solves it through reasoning. The Assistant first thinks through the reasoning process internally, then provides a clear and accurate answer. For multiple-choice questions, the Assistant must analyze each option's content, compare it with the reasoned answer, strictly select the correct answer from the given options, and specify the corresponding option (e.g., A, B, C, etc.). The reasoning process and answer are enclosed in ``<think></think>'' and ``<answer></answer>'' tags, respectively, as follows:
<think>Reasoning process, including problem analysis, solution steps, and option comparison (if applicable)</think>
<answer>Final answer</answer>

\end{tcolorbox}

\begin{tcolorbox}[colback=gray!15, colframe=black, title=Prompts for Multimodal Training and Evaluation, fonttitle=\bfseries, breakable]

A dialogue where the assistant answers a user's question. The user poses a question, and the assistant needs to resolve it. The assistant will first think through the reasoning process internally, then provide the answer, and identify the most relevant image related to the question.
The assistant has access to multiple images and must first determine which image is most relevant to the question before analyzing its content. For multiple-choice questions, the assistant must strictly select the answer from the given options and specify the corresponding option (A, B, C, D, E, etc.).
The reasoning process, answer, and image selection are enclosed in ``<think></think>'', ``<answer></answer>'', and ``<image\_selection></image\_selection>'' tags, respectively, as follows:  
<think>Reasoning process, including analysis of which image is relevant and its content</think><answer>Answer</answer><image\_selection>Image index (starting from 0)</image\_selection>

\end{tcolorbox}

\section{Instructions for Human Annotators~\label{sec:human_instruction}}

\begin{tcolorbox}[colback=gray!15, colframe=black, title=Instructions for Human Annotators, fonttitle=\bfseries, breakable]

To ensure the quality of image–question–answer triplets in our multimodal financial dataset, your task is to evaluate both the image–question alignment and the question–answer answerability.
Specifically, you will be presented with a financial image~(\emph{e.g.}, charts, tables, or visual reports), a corresponding multiple-choice question~(five options), and a reference answer generated by a large language model~(LLM) based on the textual financial report.
You are instructed to complete the following tasks:
\begin{itemize}
  \item \textbf{Assess answerability}: Determine whether the question can be reasonably answered using only the visual content of the image.
  \item \textbf{Evaluate alignment}: Verify whether the question is relevant to the image and whether the visual content supports the reference answer.
  \item \textbf{Decide on retention}: Make a binary decision on whether the image–question–answer triplet is valid and should be retained. Discard samples if the question is irrelevant, ambiguous, or lacks sufficient visual grounding.
\end{itemize}

You are explicitly instructed not to rely on external knowledge or financial expertise beyond what is observable in the image.
Your judgment should reflect a visual-only reasoning process, ensuring consistency with the multimodal design of the dataset.
\end{tcolorbox}

\section{Case Study~\label{sec:case}}

In this section, we present four our cases between LMM-R1 and FinLMM-R1 from FinData to show the effectiveness our method.

\begin{tcolorbox}[colback=gray!15, colframe=black, title=Case Study of LMM-R1 v.s. FinLMM-R1 (1), fonttitle=\bfseries, breakable]
\textbf{Title:} Performance of Shenwan Primary Industries\\
\textbf{Question\_Type:} Statistical Reasoning\\
\textbf{Question:} According to the Huaxi Golden Engineering sector rotation model, among the sectors recommended for overweight in last month, which performed the best in this month?\\
\textbf{Options:}
\begin{enumerate}[label=\Alph*.]
    \item Nonferrous Metals, Basic Chemicals
    \item Defense and Military Industry, Textiles and Apparel
    \item Transportation, Banking
    \item Nonferrous Metals, Transportation
    \item Basic Chemicals, Banking
\end{enumerate}
\textbf{Correct\_Answer:} A. 
\begin{center}
    \includegraphics[width=0.6\textwidth]{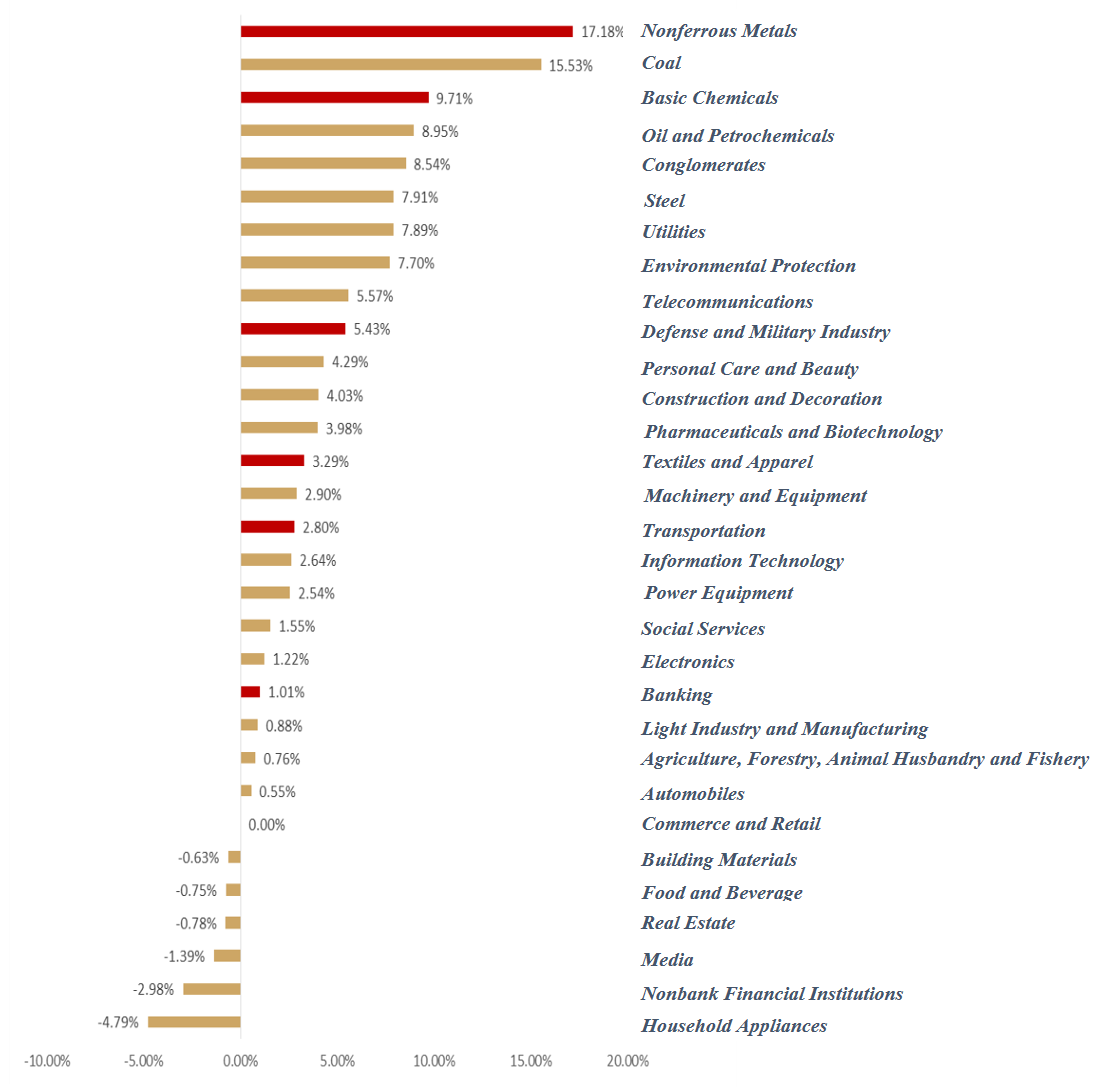}
\end{center}
\vspace{1em}
\hrule
\vspace{1em}

\textbf{LMM-R1:} \\
\textbf{<think>} \\
The user's question asks about the best-performing industries in last month as suggested by the industry rotation model, and then in this month. The relevant information for last month is provided in the first figure, where the top two performers are identified as Nonferrous Metals and Basic Chemicals. For this month, the second figure shows the best-performing indices with Nonferrous Metals leading again. Therefore, both last month and this month show that Nonferrous Metals is the best-performing industry.

The second figure also lists Transportation as another top performer in this month, which aligns with the last month recommendation. Given these points, the best-performing industries in both last month and this month are Nonferrous Metals and Transportation. Therefore, the correct answer is D.\textbf{</think>}\\
\textbf{<answer>}D. Nonferrous Metals, Transportation\textbf{</answer>}

\vspace{1em}
\hrule
\vspace{1em}

\textbf{FinLMM-R1:} \\
\textbf{<think>} \\
The relevant image for answering the question is Figure 1, which shows the performance of various indices in this month. The question asks which industries, recommended to be overweight in last month, had the best performance in this month. Looking at Figure 1, we can see that the industries with the highest positive returns in this month include Nonferrous metals and Basic chemicals. Therefore, the correct answer is A.\textbf{</think>}\\
\textbf{<answer>}A. Nonferrous metals, Basic chemicals\textbf{</answer>}

\vspace{1em}
\hrule
\vspace{1em}

\textbf{Our Analysis:} \\
This case evaluates sector-level performance through statistical reasoning.
The question involves analyzing this month's performance of Shenwan Primary Industries based on the sectors recommended for overweight last month.
Solving this problem requires cross-referencing time-sensitive investment recommendations with visualized sector performance data.
However, LMM-R1 misidentifies ``Transportation'' as a top performer, likely due to superficial keyword matching, while overlooking the actual bar heights in the chart.
This reflects limitations in both precise visual grounding and logical comparison.
In contrast, FinLMM-R1 accurately identifies ``Nonferrous Metals'' and ``Basic Chemicals'' as the top-performing industries, correctly aligning them with the prior month's recommendations.

\end{tcolorbox}

\begin{tcolorbox}[colback=gray!15, colframe=black, title=Case Study of LMM-R1 v.s. FinLMM-R1 (2), fonttitle=\bfseries, breakable]
\textbf{Title:} Quarterly Revenue Report (2020–2022)\\
\textbf{Question\_Type:} Statistical Reasoning\\
\textbf{Question:} What is the year-over-year percentage growth in the company’s most recent quarterly revenue?\\
\textbf{Options:}
\begin{enumerate}[label=\Alph*.]
    \item 8.18\%
    \item 9.18\%
    \item 16.67\%
    \item 22.39\%
    \item 29.30\%
\end{enumerate}
\textbf{Correct\_Answer:} B. 
\begin{center}
    \includegraphics[width=0.6\textwidth]{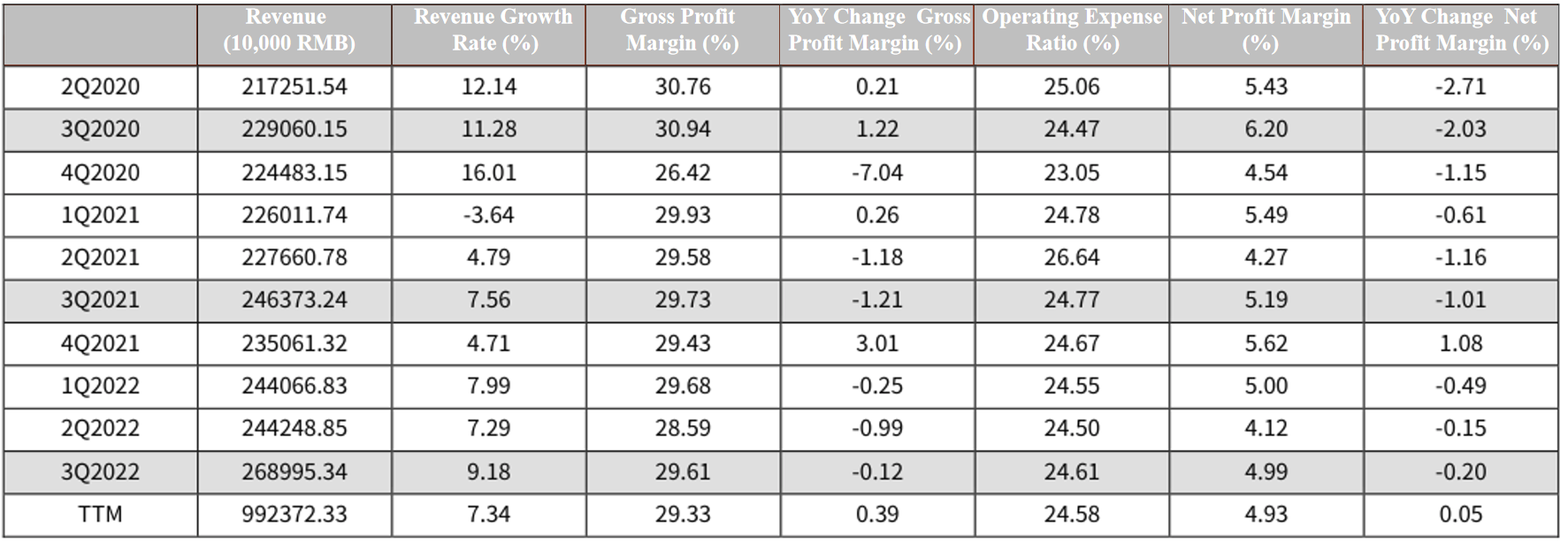}
\end{center}
\vspace{1em}
\hrule
\vspace{1em}

\textbf{LMM-R1:} \\
\textbf{<think>} \\
The user's question asks for the single-quarter Revenue Growth Rate of Hongqi Chain. We need to refer to the Revenue Growth Rate (\%) column in the provided table. This column indicates the percentage change in revenue for each quarter.
Looking at the data:
\begin{itemize}
  \item The Revenue Growth Rate for the first quarter of 2022 (1Q2022) is -1.58\%.
  \item The Revenue Growth Rate for the second quarter of 2022 (2Q2022) is 7.29\%.
\end{itemize}

The question asks for the single-quarter Revenue Growth Rate, which, in this context, refers to the highest positive growth rate during the given period. The highest positive growth rate is 7.29\%, occurring in the second quarter of 2022.
Therefore, the correct answer is D. 22.39\%.\textbf{</think>}\\
\textbf{<answer>}D. 22.39\%\textbf{</answer>}

\vspace{1em}
\hrule
\vspace{1em}

\textbf{FinLMM-R1:} \\
\textbf{<think>} \\
The relevant information for answering the question is in the second table, which provides the quarterly Revenue Growth Rates. The 3Q2022 row shows a Revenue Growth Rate of 9.18\%, which matches option B.\textbf{</think>}\\
\textbf{<answer>}B. 9.18\%\textbf{</answer>}

\vspace{1em}
\hrule
\vspace{1em}

\textbf{Our Analysis:}

This case evaluates the ability of the LMM to interpret and extract the correct quarterly revenue growth rate from a financial report.
The question asks for the year-over-year percentage growth in the company’s most recent quarterly revenue.
Solving this task requires identifying the correct time period (3Q2022) and accurately locating the corresponding value in the ``Revenue Growth Rate (\%)'' column of the table.
LMM-R1 misunderstands the question and instead selects the highest value from the entire column (22.39\%), which corresponds to 2Q2022 rather than the most recent quarter.
This mistake reflects its limited understanding of temporal references and the intended question focus.
In contrast, FinLMM-R1 accurately identifies 3Q2022 as the most recent quarter, extracts the correct growth rate of 9.18\%, and selects the correct answer, showcasing stronger alignment with financial context and temporal reasoning.

\end{tcolorbox}

\begin{tcolorbox}[colback=gray!15, colframe=black, title=Case Study of LMM-R1 v.s. FinLMM-R1 (3), fonttitle=\bfseries, breakable]
\textbf{Title:} 2021 Annual Results (Year Ended December 31, in RMB millions)\\
\textbf{Question\_Type:} Arithmetic Reasoning\\
\textbf{Question:} What was the year-over-year percentage increase in the company’s net profit attributable to shareholders in 2021?\\
\textbf{Options:}
\begin{enumerate}[label=\Alph*.]
    \item 40.8\%
    \item 45.5\%
    \item 78.0\%
    \item 65\%
    \item 75\%
\end{enumerate}
\textbf{Correct\_Answer:} C. 
\begin{center}
    \includegraphics[width=0.6\textwidth]{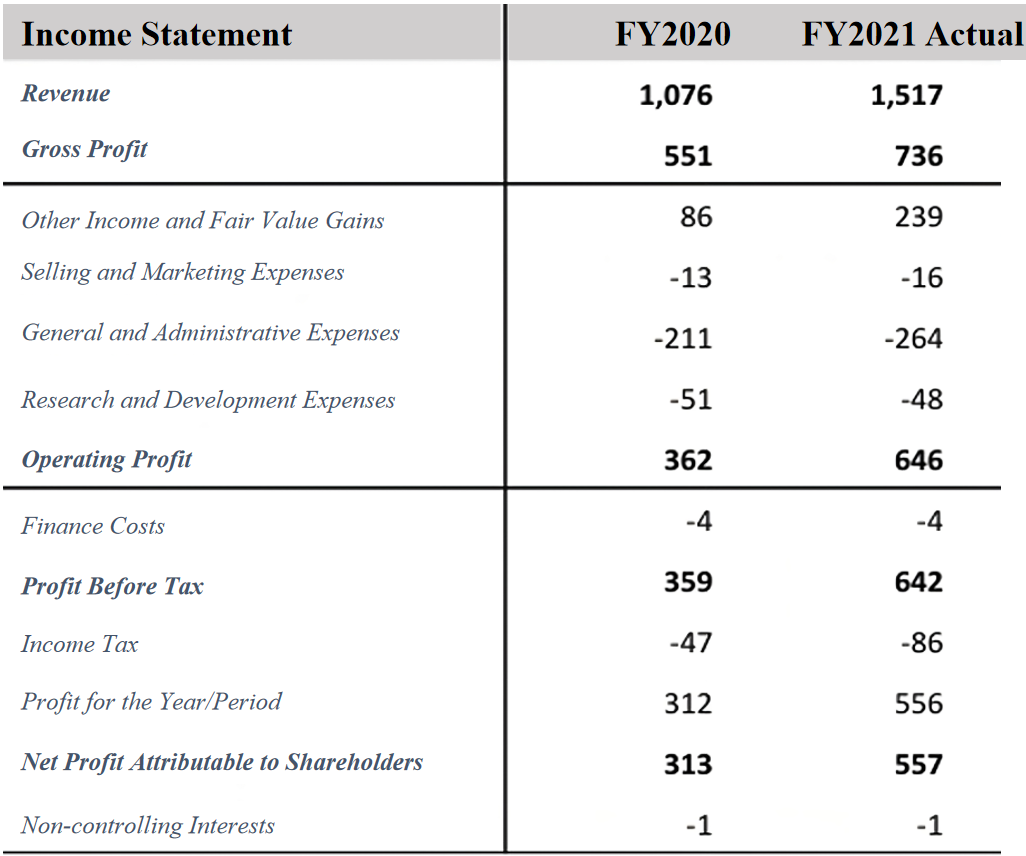}
\end{center}
\vspace{1em}
\hrule
\vspace{1em}

\textbf{LMM-R1:} \\
\textbf{<think>} \\
The question asks for the percentage increase in the company's Net Profit Attributable to Shareholders from 2020 to 2021. Looking at the Net Profit Attributable to Shareholders row in the 2021 table, we can see the values as follows:
2020: 313 million RMB; 2021: 557 million RMB.
To find the percentage increase, we use the formula:\[
\text{Percentage Increase} = \left( \frac{\text{New Value} - \text{Old Value}}{\text{Old Value}} \right) \times 100
\]
Substituting the values:\[
\text{Percentage Increase} = \left( \frac{557 - 313}{313} \right) \times 100
\]
\[
\text{Percentage Increase} = \left( \frac{244}{313} \right) \times 100
\]
\[
\text{Percentage Increase} = 0.75 \times 100
\]
\[
\text{Percentage Increase} = 75\%
\]
Therefore, the correct answer is E. 75\%.\textbf{</think>}\\
\textbf{<answer>}E. 75\%\textbf{</answer>}

\vspace{1em}
\hrule
\vspace{1em}

\textbf{FinLMM-R1:} \\
\textbf{<think>} \\
The relevant information for answering the question is in Table 1, which shows the financial results of the company. The Net Profit Attributable to Shareholders row indicates the shareholder net profit for both 2020 and 2021. The percentage increase from 2020 to 2021 is calculated as follows:\[
\text{Percentage Increase} = \left( \frac{\text{Net Profit}_{2021} - \text{Net Profit}_{2020}}{\text{Net Profit}_{2020}} \right) \times 100
\]
From the table:\[
\text{Net Profit}_{2021} = 557\ \text{million RMB}
\]
\[
\text{Net Profit}_{2020} = 313\ \text{million RMB}
\]
\[
\text{Percentage Increase} = \left( \frac{557 - 313}{313} \right) \times 100 = \left( \frac{244}{313} \right) \times 100 \approx 77.96\%
\]Therefore, the correct answer is C. 78.0\%.
\textbf{</think>}\\
\textbf{<answer>}C. 78.0\%\textbf{</answer>}

\vspace{1em}
\hrule
\vspace{1em}

\textbf{Our Analysis:} \\
This case presents an example assessing arithmetic reasoning based on an annual financial report.
The question requires computing the year-over-year percentage increase in net profit attributable to shareholders from 2020 to 2021.
Solving this task involves precise numerical extraction from tabular financial data and application of a standard percentage growth formula.
LMM-R1 makes a computational error: it correctly identifies the net profit figures but miscalculates the final percentage, arriving at 75\% instead of the accurate 78.0\%.
This exposes its limitations in numerical precision and basic arithmetic reasoning.
In contrast, FinLMM-R1 correctly extracts the relevant data, applies the formula, and arrives at the accurate percentage increase of 78.0\%, demonstrating superior capability in financial arithmetic reasoning and table-based numerical understanding.

\end{tcolorbox}

\begin{tcolorbox}[colback=gray!15, colframe=black, title=Case Study of LMM-R1 v.s. FinLMM-R1 (4), fonttitle=\bfseries, breakable]
\textbf{Title:} Stock Options Volume PCR\\
\textbf{Question\_Type:} Financial Knowledge\\
\textbf{Question:} What does PCR represent in the options market?\\
\textbf{Options:}
\begin{enumerate}[label=\Alph*.]
    \item Price Change Rate
    \item Open Interest Ratio
    \item Trading Volume Ratio
    \item Volatility Ratio
    \item Open Contracts Ratio
\end{enumerate}
\textbf{Correct\_Answer:} C. 
\begin{center}
    \includegraphics[width=0.6\textwidth]{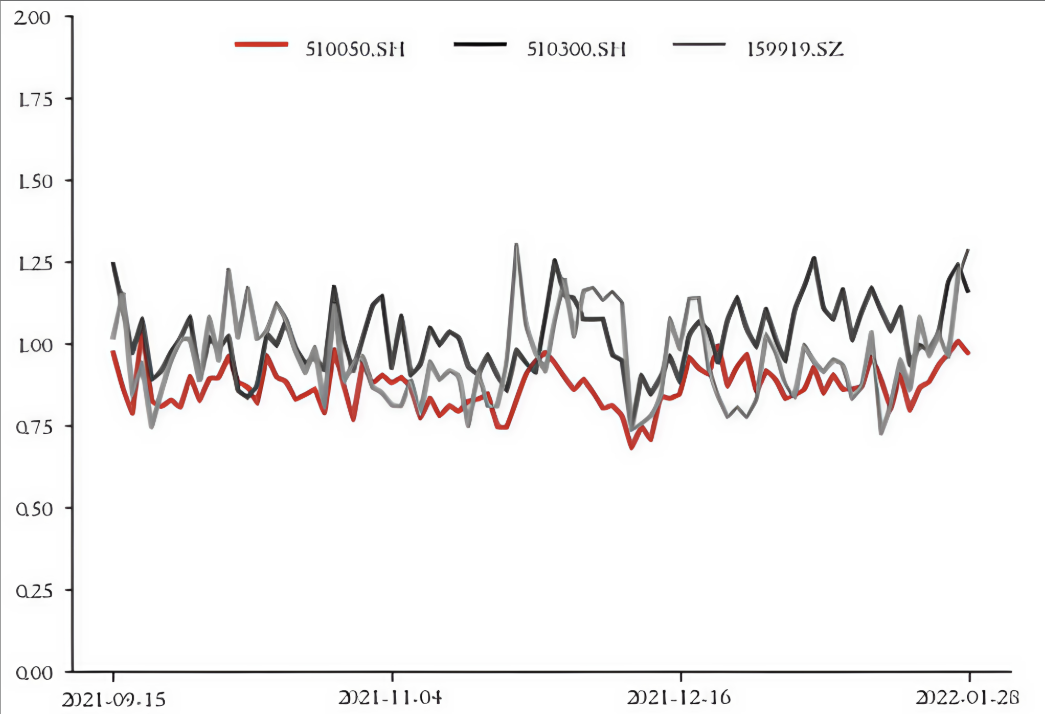}
\end{center}
\vspace{1em}
\hrule
\vspace{1em}

\textbf{LMM-R1:} \\
\textbf{<think>} \\
The question asks about the meaning of PCR in the context of options trading. PCR stands for "Put-Call Ratio," which is a measure used in options trading to gauge the relative preference of traders for buying puts versus calls. This ratio helps in understanding the market sentiment towards call options relative to put options.
Looking at the images provided, none of them directly show PCR. However, the second image (Figure 2) shows the stock option trading volume and the PCR (Put-Call Ratio). This indicates that PCR is the ratio of the volume of put options to the volume of call options.
Therefore, the correct answer is: B. 
\textbf{</think>}\\
\textbf{<answer>}B. Open Interest Ratio\%\textbf{</answer>}

\vspace{1em}
\hrule
\vspace{1em}

\textbf{FinLMM-R1:} \\
\textbf{<think>} \\
PCR in the options market stands for Put-Call Ratio, which is a sentiment indicator measuring the relative trading activity between put and call options. It is typically defined as the ratio of put option volume to call option volume:

\[
\text{PCR} = \frac{\text{Put Option Volume}}{\text{Call Option Volume}}
\]

This metric reflects market sentiment — a high PCR may indicate bearish sentiment, while a low PCR suggests bullishness. PCR is not related to price change rate, volatility ratio, or open interest ratio. Although there is a version based on open interest, unless otherwise specified, PCR usually refers to volume ratio. 
Therefore, the correct answer is C. Trading Volume Ratio.\textbf{</think>}\\
\textbf{<answer>}C. Trading Volume Ratio\textbf{</answer>}

\vspace{1em}
\hrule
\vspace{1em}

\textbf{Our Analysis:}

This case evaluates the ability of the LMM to interpret financial terminology, specifically the meaning of ``PCR'' in the context of the options market.
The question asks what PCR stands for, and correctly answering it requires financial domain knowledge, specifically, recognizing that PCR (Put–Call Ratio) typically refers to the trading volume ratio of put options to call options, unless otherwise specified.
LMM-R1 misinterprets the question by associating PCR with open interest rather than trading volume.
While an alternative definition of PCR based on open interest exists, it is not the default interpretation in most financial contexts.
As a result, LMM-R1 incorrectly selects Option B~(Open Interest Ratio), reflecting limited sensitivity to default financial conventions and a lack of disambiguation ability when no clarifying context is provided.
FinLMM-R1 correctly understands that PCR commonly refers to the trading volume ratio, identifies the definition accordingly, and selects Option C~(Trading Volume Ratio).
This demonstrates stronger alignment with financial domain standards and superior ability to resolve potential ambiguity using contextual cues.

\end{tcolorbox}

\section{Statistical Significant Test\label{sec:stat}}

We perform the significance test between our FinLMM-R1 and LMM-R1~\cite{peng2025lmm}. 
We run our FinLMM-R1 and LMM-R1 10 times on MMStar and MathVista with random seeds ranging from 1 to 10. 
The accuracy of our FinLMM-R1 are 61.14\(\pm\)0.92 and 65.29\(\pm\)1.02 in MMStar and MathVista dataset, while the accuracy of LMM-R1 are 58.13\(\pm\)0.78 and 63.31\(\pm\)1.21 in MMStar and MathVista dataset. 
At the significance level 0.05, we perform significance test to verify that our FinLMM-R1 is better than LMM-R1. 
The p-values in MMStar and MathVista dataset are 2.97$\times$10$^{-7}$ and 9.25$\times$10$^{-4}$, which is far below 0.05, demonstrating the superiority of our FinLMM-R1 is statistically significant.

\section{Limitation and Future Direction\label{sec:lim}}

While our framework achieves marked improvements in both general and financial multimodal reasoning, its adversarial‐reward module currently relies on a BERT-based discriminator due to practical compute constraints.
This choice may limit the depth of reasoning signals that a more powerful model could capture. 
In future work, we plan to integrate larger pretrained architectures (\emph{e.g.}, 3B or 7B LMMs) as discriminators to guide the internal ``thinking'' process. 
Additionally, our data–collection and annotation pipeline has been thoroughly validated in the financial domain. 
In the future, we intend to adapt and generalize our annotation workflow across diverse domains (such as healthcare, legal, and scientific multimedia) and develop a unified, domain-agnostic reasoning dataset that can further strengthen model robustness and cross‐domain transfer.

\section{Broader Impacts\label{sec:imp}}
By innovating on reward design for multimodal reasoning, our work paves the way for more robust test-time scaling of reasoning models: richer feedback signals can accelerate adoption of scalable inference strategies across vision, language, and beyond. 
Moreover, our domain-tailored PDF annotation pipeline, demonstrated on financial research reports, offers a blueprint for automated data extraction from professional documents, potentially unlocking large, diverse corpora (\emph{e.g.}, legal, scientific) and thereby broadening the reach and applicability of multimodal reasoning systems.

\end{document}